\documentclass[11pt,letterpaper]{article}

\usepackage[T1]{fontenc}
\usepackage[utf8]{inputenc}
\usepackage{lmodern}
\usepackage[margin=1in]{geometry}
\usepackage{microtype}
\usepackage{amsmath,amssymb}
\usepackage{textcomp}
\usepackage{booktabs,longtable,array}
\usepackage[table]{xcolor}
\usepackage{pdflscape}
\usepackage{enumitem}
\usepackage{caption}
\usepackage{authblk}
\usepackage{titlesec}
\usepackage[hang]{footmisc}
\usepackage[hyphens]{url}
\usepackage{xurl}
\usepackage[colorlinks=true,linkcolor=linkblue,urlcolor=linkblue,citecolor=linkblue]{hyperref}
\usepackage{bookmark}

\definecolor{linkblue}{RGB}{31,78,121}
\definecolor{groupgray}{gray}{0.92}
\setlist{itemsep=2pt,topsep=4pt}

\titleformat{\section}{\Large\bfseries}{\thesection.}{0.6em}{}
\titleformat{\subsection}{\large\bfseries}{\thesubsection.}{0.6em}{}
\titleformat{\subsubsection}{\normalsize\bfseries}{\thesubsubsection.}{0.6em}{}
\hypersetup{
  pdftitle={Key Institutional Considerations for Designing Effective Independent Verification Organizations},
  pdfauthor={Uma Kalkar, Rumman Chowdhury, Reva Schwartz, Gabriella Waters}
}

\title{\textbf{Key Institutional Considerations for Designing Effective\\Independent Verification Organizations}}
\author[1,2]{Uma Kalkar}
\author[1,2]{Rumman Chowdhury}
\author[1,3]{Reva Schwartz}
\author[1,3,4]{Gabriella Waters}
\affil[1]{Independent AI Evaluation Foundation}
\affil[2]{Humane Intelligence PBC}
\affil[3]{Civitaas Insights}
\affil[4]{Virginia State University}
\date{}

\begin{document}
\maketitle

\begin{abstract}
\noindent Independent verification organizations (IVOs) have emerged as a popular,
bipartisan solution for facilitating high-quality independent
evaluations of AI technologies. But momentum is outpacing design: there
is emerging consensus on the necessary operating conditions for robust
IVOs, but little discourse on institutional and policy specifications to
ensure those conditions are met. IVOs are currently envisioned primarily
as a mechanism operating at the AI frontier and model level to evaluate
catastrophic risks. Little work has yet been done on how to establish or
enforce an IVO ecosystem that is equipped to handle myriad risks (i.e.,
beyond catastrophic ones) and operate at the system and application
level across the deployment and use stages of the AI lifecycle.
\textbf{This paper argues that IVO design should ensure the capacity for
evaluating AI models, systems, and applications in context for both
frontier companies and enterprise companies \& AI-adopting
organizations. Additionally, this paper provides early considerations to
advance IVO discourse from normative to operative, and from existential
to everyday risks.} Drawing from existing institutional constructs in
financial services, education, public health, and science, we outline
key considerations on whom IVOs serve, what access they require, how
they interact with external governing structures, how they are funded,
and how they fit into the wider assurance ecosystem. Much of the
structure and knowledge necessary to holistically design IVOs and their
ecosystem does not yet exist, even as companies rush to ink agreements
with their chosen evaluators.
\end{abstract}

\clearpage
\tableofcontents
\clearpage

\section{Introduction}\label{introduction}

The AI industry has operated with few constraints: companies race each
other to market, and when obliged to test products for safety, they set
their own rules and grade their own homework. For years, safety measures
have been pitted against deployment, treated as a hindrance to getting
models to market rather than a condition for doing so
(\href{https://www.ft.com/content/8253b66e-ade7-4d1f-993b-2d0779c7e7d8}{Criddle
2025};
\href{https://www.forbes.com/councils/forbestechcouncil/2026/10/14/why-ai-evaluation-needs-independence-not-just-human-judgment/}{Majumdar
2026}). In recent months, a rash of agentic AI breaches -- illegally
hacking into government data
(\href{https://www.nytimes.com/2026/09/25/world/australia/openai-hack-australia.html}{Kim
2026};
\href{https://www.nytimes.com/2026/09/25/technology/openais-ai-us-government-websites.html}{Conger,
Swanson, and Kang 2026}), swarming into external repositories
(\href{https://www.nytimes.com/2026/07/21/technology/openai-attack-hugging-face.html}{Conger
2026}), and leaking user information
(\href{https://www.reuters.com/world/openai-works-understand-full-scope-agent-activity-user-data-leak-emerges-2026-09-25/}{Seetharaman,
Satter, and Horwitz 2026}) -- has ignited calls for a more mature
assurance ecosystem comprising better measurement, evaluation, and
reporting on AI tools in use. Independent verification organizations
(IVOs) offer a potential answer.

\subsection{Problem Statement}\label{problem-statement}

As currently envisioned, an IVO is a licensed, independent, expert-led
body that impartially assesses whether an AI system meets specific
safety, risk, and trustworthiness criteria
(\href{https://fathom.cdn.prismic.io/fathom/aULIU3NYClf9oW4F_fathom-ivo-report.pdf}{Fathom
2025};
\href{https://dls.virginia.gov/commissions/jcots/Materials/2026_Independent\%20Verification\%20Organizations\%20and\%20AI\%20Oversight\%20in\%20Virginia.pdf}{Smith,
McKenzie, and Jones 2026}).\footnote{It is worth noting that these
  criteria have yet to be fully established or harmonized across state
  and federal levels.} The market for IVOs is forming quickly and has
drawn bipartisan support. But momentum is outpacing design. There is
emerging consensus on the necessary operating conditions for robust IVOs
(\href{https://www.aef.one/aef-one.pdf}{Stosz et al. 2025});
however, IVOs are currently envisioned primarily as a mechanism
operating at the AI frontier and model level to evaluate catastrophic
risks. Little work has yet been done on how to establish or enforce an
IVO ecosystem that is equipped to handle myriad risks (i.e., beyond
catastrophic ones) and operate at the system and application level
across the deployment and use stages of the AI lifecycle. \textbf{This
paper argues that IVO design should ensure the capacity for evaluating
AI models, systems, and applications in context for both frontier
companies and enterprise companies \& AI-adopting
organizations.}\footnote{This paper distinguishes between two types of
  companies \& organizations engaging with IVOs. \textbf{Frontier
  companies} build and train general-purpose AI that they develop and
  provide to others. \textbf{Enterprise companies \& AI-adopting
  organizations} deploy, adopt, implement, and use AI tools developed by
  others within their own sectors.} Mandates without sound institutional
design can leave oversight vulnerable to regulatory capture. Without
clear and appropriate constructs guiding IVO design, this field risks
succumbing to said capture. Good institutional design also reduces
wasted resources while driving programmatic efficiency by clarifying
responsibilities, avoiding unnecessary duplication, and accounting for
evidence needs across the ecosystem from the outset, rather than
retrofitting methods and institutions to serve overlooked constituencies
later. To this end, this paper provides early considerations to advance
IVO discourse from normative to operative, and from existential to
everyday risks.

\subsection{Research Questions}\label{research-questions}

This paper covers five research questions (RQs) that repeatedly emerge
regarding IVO design, conduct, and engagement:

\begin{itemize}
\item
  RQ1: Whom do IVO evaluations serve?
\item
  RQ2: How can IVO mandates be designed to ensure adequate and
  appropriate access that enables oversight while protecting proprietary
  information?
\item
  RQ3: How can a successful IVO regime be developed while appreciating
  the nascency of the evaluator workforce and science of evaluation?
\item
  RQ4: How can IVOs be financially independent of the entities they are
  evaluating, without being a taxpayer burden?
\item
  RQ5: Which functions belong to IVOs, and which belong to other bodies
  in the assurance ecosystem?
\end{itemize}

\subsection{Who This Paper Is For}\label{who-this-paper-is-for}

Considerations are provided for the following audiences (see Section 9
for takeaways):

\begin{itemize}
\item
  \emph{Regulators and legislators:} on what to require of IVOs when
  drafting regulation and legislation for IVO and IVO ecosystem
  governance;
\item
  \emph{Oversight and intermediary bodies (e.g., accreditors,
  standards-setting bodies, funding intermediaries):} on how to
  structure the operational relationships between frontier companies,
  enterprise companies \& AI-adopting organizations, IVOs, and
  organizations commissioning validation or other oversight checks;
\item
  \emph{Organizations considering becoming IVOs:} on what operating and
  conduct requirements they should expect to meet; and
\item
  \emph{Frontier companies and enterprise companies \& AI-adopting
  organizations -- the two distinct markets engaging with IVOs:} on what
  to expect IVOs to ask of them during assessment.
\end{itemize}

Throughout, this paper distinguishes between requirements for
\textbf{IVO designation \& oversight bodies}\footnote{The \textbf{IVO
  designation \& oversight body} decides which organizations qualify as
  IVOs and checks that they keep meeting that bar. The body can belong
  to a government's regulatory agency or be a standalone institution.
  See Section 4.2 for more information about the body's mandate and
  Section 5.4 for more information on the body's makeup.} and
\textbf{IVOs themselves}. Recognizing that an IVO may range from an
established firm to a small- to medium-sized business to an individual
consultancy, the minimum viable obligations for organizations across
this spectrum are outlined.

\subsection{Paper Outline}\label{paper-outline}

\begin{itemize}
\item
  Section 1 (Introduction) outlines the problem, research questions, and
  audiences covered by this paper.
\item
  Section 2 (Background) covers the current state of AI evaluation.
\item
  Section 3 (RQ1) maps the IVO model and the current U.S. state and
  federal legislative landscape.
\item
  Section 4 (RQ2) details what level of information should be required
  from companies to ensure IVOs have adequate and appropriate access.
\item
  Section 5 (RQ3) proposes how evaluator expertise and IVO designation,
  standards, and validation methods can best develop alongside
  evaluation science.
\item
  Section 6 (RQ4) considers funding models to preserve IVO integrity and
  independence.
\item
  Section 7 (RQ5) specifies which assurance functions belong to IVOs and
  which belong to other bodies.
\item
  Section 8 (Further Research) identifies gaps and next steps for
  continued scholarship.
\item
  Section 9 (Recommendations Across Audiences) distills key
  considerations for each audience (Section 1.3) and concludes.
\end{itemize}

\section{Background}\label{background}

\subsection{Current State of AI
Evaluation}\label{current-state-of-ai-evaluation}

AI evaluation itself is not new: indeed, the Turing Test can be
classified as the first deception evaluation. The methods for evaluating
AI technologies are still maturing. See
\href{https://dl.acm.org/doi/pdf/10.1145/3491209}{Kaur et al.
(2022)} for an overview of trustworthy evaluation, validation, and
verification methods;
\href{https://arxiv.org/html/2502.15620v1\#bib.bib13}{Burden et al.
(2025)} for analysis of common evaluation paradigms;
\href{https://eprints.whiterose.ac.uk/id/eprint/189769/1/paper3.pdf}{Cohn
et al. (2022}) for a framework to categorize AI evaluation instruments;
and \href{https://arxiv.org/pdf/2307.03109}{Chang et al. (2023)}
for an overview of evaluation methods for large language models (LLMs).

Evaluation helps us understand how decisions by AI models, systems, and
applications (hereafter referred to broadly as `AI tools') are reached
(\href{https://alumni.media.mit.edu/~kris/ftp/EGPAI_2016_paper_9.pdf}{Bieger
et al. 2016}), whether those decisions meet ethical standards
(\href{https://link.springer.com/article/10.1007/s43681-024-00504-8}{Waters,
Mapp, and Honenberger 2025}), and what impacts they have, from
``first-order'' direct outputs to ``second-order'' effects on the
broader society (\href{https://arxiv.org/pdf/2505.18893}{Schwartz et
al. 2025}). These insights allow us to check an AI tool for harmful
behavior and gauge its capability, reliability, and trustworthiness
(\href{https://dl.acm.org/doi/pdf/10.1145/3491209}{Kaur et al.
2022}). Current literature suggests spotty coverage of the breadth of
harms caused by AI tools in the contexts and modalities within which
they are used. \href{https://arxiv.org/pdf/2310.11986}{Weidinger et
al. (2023)} point to an evaluation gap, given that current evaluations
focus primarily on assessing text models' capabilities, leaving
higher-order human interaction and systemic impacts underassessed.
\href{https://arxiv.org/pdf/2607.14353v1}{Kroll et al. (2026)}
argue that safety arises from interactions within the wider
sociotechnical system and cannot be inferred from model performance
alone. They also note that emerging evaluation standards largely address
model behavior without considering system-level interactions or
outcomes.

Evaluations are time- and resource-intensive and require privileged
information -- data, logs, and model access. Importantly, privileged
information is primarily held and protected by \textbf{frontier
companies}, which develop and deploy AI tools. Because of this dynamic,
frontier companies largely control and dictate the evaluation process,
including what gets assessed, who conducts the tests, who pays, what is
released publicly, and how results are interpreted. Implications of this
arrangement extend beyond the technology sector to \textbf{enterprise
companies \& AI-adopting organizations} across the economy, which deploy
and use AI tools built by others for a range of uses, purposes, and
markets. The application of AI tools to these contexts calls for
\emph{in situ} evaluations, yet these companies rarely have access to
the developer information needed to conduct them.

\subsection{Current State of Independent AI
Evaluation}\label{current-state-of-independent-ai-evaluation}

Securing evaluator \emph{independence} is a foundational institutional
design challenge. External evaluator engagements typically happen when
frontier companies opt in for them
(\href{https://www.hsgac.senate.gov/wp-content/uploads/Chris-Painter-Testimony.pdf}{Painter
2026}) or when they commission and pay for them
(\href{https://openai.com/index/strengthening-safety-with-external-testing/}{OpenAI
2025};
\href{https://www.anthropic.com/news/accenture-embedded-evaluation}{Anthropic
2026a}). Either way, it is the frontier company side of the arrangement
that controls the terms: the former does not create incentives for
independent evaluation and can still have reporting constraints; the
latter can create conflicts of interest
(\href{https://dl.acm.org/doi/abs/10.5555/3716662.3716784}{Stein et
al. 2024}). Invited and fully external third-party evaluators have
faced significant financial, capacity, and information challenges.
Developers have been reported to limit evaluator access to model details
(\href{https://arxiv.org/pdf/2609.25413}{Charnock et al. 2026}) and
surrounding information about the model's societal impact
(\href{https://arxiv.org/html/2511.05613v1}{Reuel et al. 2025}), as
well as demand razor-thin evaluation turnaround timelines
(\href{https://www.aef.one/aef-one.pdf}{Stosz et al. 2025}) that
create barriers to providing confident external assessment.

In September 2026, in response to mounting public and political
pressure, Anthropic and OpenAI committed to ``embedding'' independent
evaluators with ``employee-like'' privileges inside their labs to
provide access to models before deployment
(\href{https://darioamodei.com/post/we-must-pace-the-frontier}{Amodei
2026};
\href{https://techcrunch.com/2026/09/16/anthropic-and-openai-want-to-embed-safety-evaluators-will-they-really-be-independent/}{Bellan
2026}). Anthropic then partnered with Accenture to spin out a
full-service evaluation practice
(\href{https://www.anthropic.com/news/accenture-embedded-evaluation}{Anthropic
2026a}). Yet already, tensions over publishing findings threaten the
independence of these evaluations
(\href{https://www.reuters.com/world/behind-ai-industrys-call-independent-reviews-fight-over-what-gets-made-public-2026-10-13/}{Soni
2026}). Constraints on information access and the time available for
completing evaluations also raise questions about the quality of methods
used and the strength of the resulting findings
(\href{https://techcrunch.com/2026/09/16/anthropic-and-openai-want-to-embed-safety-evaluators-will-they-really-be-independent/}{Bellan
2026}).

Today, the practice of independent evaluation relies on the goodwill of
a small group of high-value frontier companies. Surrounding governing
structures, such as standards-setting organizations, accreditation hubs,
and adjudication bodies, are nascent. This exposes independent
evaluation and the IVO ecosystem as a whole to industry and ideological
capture. The seeds of this are already taking root as IVOs are
increasingly framed as ``embedded'' assessors working inside frontier
companies (\href{https://arxiv.org/pdf/2609.25413}{Charnock et al.
2026}). Normalization of evaluation conducted on frontier company terms
reinforces a narrow, developer-centered purpose for independent
evaluation. Policymakers should respond by drawing on established
institutional models to cultivate an evaluation ecosystem that is
sustainable and credible, and that serves organizations across sectors
as well as the broader public.

\section[RQ1: Whom IVO evaluations serve]{RQ1: Whom do IVO evaluations
serve?}\label{rq1-whom-do-ivo-evaluations-serve}

\subsection{The IVO Model}\label{the-ivo-model}

An IVO is envisioned as a licensed, independent, expert-led body that
impartially assesses whether an AI system meets specific safety, risk,
and trustworthiness criteria
(\href{https://fathom.cdn.prismic.io/fathom/aULIU3NYClf9oW4F_fathom-ivo-report.pdf}{Fathom
2025};
\href{https://dls.virginia.gov/commissions/jcots/Materials/2026_Independent\%20Verification\%20Organizations\%20and\%20AI\%20Oversight\%20in\%20Virginia.pdf}{Smith,
McKenzie, and Jones 2026}).

An IVO's role should be to verify whether an AI tool meets specified
requirements under an established framework and within a defined
evaluation boundary. Evaluators should review the relevant evidence and
conduct or supervise authorized testing to determine whether the AI tool
meets the specified requirements and to identify AI tool limitations.
They should then issue findings, whether favorable or unfavorable, and
document the basis for each. A finding should apply only within that
evaluation boundary and to the uses, conditions, and limitations
specified in the verification record, which include, but are not limited
to:

\begin{itemize}
\item
  The system or service being assessed, including the relevant version,
  components, and material integrations;
\item
  The stated purpose, permitted uses, excluded uses, and operating
  environment;
\item
  The users, operators, potentially affected people, and allocation of
  human decision authority;
\item
  The data inputs, outputs, external tools, and systems that materially
  influence performance or outcomes;
\item
  The controls, oversight processes, and constraints that are within the
  scope of the finding; and,
\item
  The elements excluded from review and any limitations those exclusions
  place on the conclusion.
\end{itemize}

IVO evaluation should attach to a defined AI-enabled service or decision
process. Its findings should not be generally applied to an AI
developer, model family, or product brand in the abstract. For instance,
an evaluation conducted on a specific version of an LLM will need to be
reviewed and reassessed if and when an updated version of the LLM is
deployed. The same should apply to any transformative or derivative
changes to the technology stack. Further, a finding from one evaluation
boundary should not transfer automatically to a different deployment
setting, workflow, customer base, jurisdiction, user population, or
consequential use. This restriction is needed because a system's risk
profile and performance can change with its configuration, operating
context, and institutional use.

Verification can attach at several moments in the AI lifecycle: a model
before release, a product built on that model, a deployment in a
specific setting, and the system once in use. Because point-in-time
checks go stale as systems change, verification should continue after
deployment, drawing on monitoring of incidents and real-world use
(\href{https://arxiv.org/abs/2410.04931}{Stein, Bernardi, and Dunlop
2024}).

Not every AI tool requires independent verification. Lawmakers should
establish a risk-based remit for IVOs, specifying the criteria for
determining which systems require independent evaluation, including the
severity and reversibility of potential harm, use of sensitive data,
degree of autonomy, scale or vulnerability of affected people,
contestability of consequential decisions, and the extent to which
existing oversight can independently assess the system. IVOs should then
apply the established criteria consistently within their authorized
remit. To assuage concerns about ideological capture or normative risk
prioritization, IVOs should not independently determine which risks
warrant regulatory intervention (Section 7).

Before engaging an IVO, two questions must be answered: (1) does the AI
tool in its specific use case require evaluation? and (2) is an IVO
required to conduct the evaluation? Evaluation must be proportionate to
the risks posed in use. Sometimes, developers' internal testing and
assessment or demonstration of compliance with existing sectoral rules
can be enough. Crucially, the need for an IVO depends on the
\emph{context of deployment} rather than on the model on its own. For
example, the same AI model may be used to screen job applicants and
draft emails, but only the former may warrant an IVO evaluation.
Respondents in a survey conducted by
\href{https://dls.virginia.gov/commissions/jcots/Materials/2026_Independent\%20Verification\%20Organizations\%20and\%20AI\%20Oversight\%20in\%20Virginia.pdf}{Smith,
McKenzie, and Jones (2026)} found IVOs useful primarily for high-risk
AI deployment and uses, where potential harm is ``severe or hard to
reverse,'' or where existing oversight and capacity are insufficient.
They also rated a hybrid model that combines IVOs with other mechanisms
``at least as favorably'' as IVOs alone for catching real harms.

\subsection{Legislation Around
IVOs}\label{legislation-around-ivos}

IVOs and independent evaluations have gained support at the state and
federal levels amid growing public demand. Nine out of ten Americans
support guardrails for AI systems
(\href{https://poll.qu.edu/poll-release?releaseid=3969}{Quinnipiac
University 2026}), with 56 percent of surveyed Americans entrusting the
development of such guardrails to ``independent subject matter experts''
(\href{https://fathom.cdn.prismic.io/fathom/aULIU3NYClf9oW4F_fathom-ivo-report.pdf}{Fathom
2025}). Eighty-six percent of Americans are in favor of mandating
``independent safety standards'' for AI developers, even at the expense
of slower AI development
(\href{https://poll.qu.edu/poll-release?releaseid=3969}{Quinnipiac
University 2026}). On IVOs, 67 percent have a positive impression of
the concept, and 70 percent believe that IVOs are ``an important first
step'' for AI safety governance
(\href{https://fathom.cdn.prismic.io/fathom/aULIU3NYClf9oW4F_fathom-ivo-report.pdf}{Fathom
2025}).

Legislators have responded with a mix of enacted and proposed
approaches:

\begin{itemize}
\item
  \emph{Enacted Frameworks:}

  \begin{itemize}
  \item
    In September 2026, California passed a first-in-the-nation framework
    to classify and regulate IVOs
    (\href{https://leginfo.legislature.ca.gov/faces/billTextClient.xhtml?bill_id=202520260SB813}{SB
    813}). Under SB 813, organizations can earn IVO designation from
    the state's Government Operations Agency by demonstrating
    ``expertise, independence, and documented methods for {[}AI{]}
    evaluation''
    (\href{https://leginfo.legislature.ca.gov/faces/billTextClient.xhtml?bill_id=202520260SB813}{SB
    813}). While the framework mandates that would-be IVOs receive the
    designation by 2028, it does \emph{not} require developers,
    deployers, or operators of AI tools to obtain an IVO evaluation
    (\href{https://leginfo.legislature.ca.gov/faces/billTextClient.xhtml?bill_id=202520260SB813}{SB
    813}). A companion bill,
    \href{https://calmatters.digitaldemocracy.org/bills/ca_202520260ab1405}{AB
    1405}, requires AI auditors to register with the state by 2029.
  \item
    Connecticut's
    \href{https://www.cga.ct.gov/asp/CGABillStatus/cgabillstatus.asp?selBillType=Bill\&bill_num=HB5222}{HB
    5222}, an omnibus bill passed in May 2026, launched a pilot IVO
    program through which the state can select up to five expert-led
    IVOs to assess AI systems against state-set risk standards
    (\href{https://www.cga.ct.gov/asp/CGABillStatus/cgabillstatus.asp?selBillType=Bill\&bill_num=HB5222}{HB
    5222};
    \href{https://fathom.org/insights/pr-newswire-fathom-applauds-connecticut-passing-legislation-that-establishes-independent-verification-(ivo)-for-ai-governance}{Fathom
    2026}).
  \item
    Illinois's
    \href{https://www.ilga.gov/Legislation/BillStatus/FullText?LegDocId=197587\&DocName=10400SB0315enr\&DocNum=315\&DocTypeID=SB\&LegID=157797\&GAID=18\&SessionID=114\&SpecSess=\&Session=}{SB
    315}, codified in July 2026, mandates that ``large frontier AI
    developers'' undergo annual independent third-party audits conducted
    by qualified professionals for ``catastrophic risk assessments''
    starting in 2028
    (\href{https://www.ilga.gov/Legislation/BillStatus/FullText?LegDocId=197587\&DocName=10400SB0315enr\&DocNum=315\&DocTypeID=SB\&LegID=157797\&GAID=18\&SessionID=114\&SpecSess=\&Session=}{SB
    315}).
  \item
    Virginia's
    \href{https://lis.virginia.gov/bill-details/20261/HB797}{HB
    797}, enacted in April 2026, enables the state's Technology and
    Science Commission to assess whether IVOs would be a feasible and
    useful AI governance mechanism for the state
    (\href{https://lis.virginia.gov/bill-details/20261/HB797}{HB
    797}).
  \end{itemize}
\item
  \emph{Proposed Frameworks:}

  \begin{itemize}
  \item
    Ohio introduced a bill in December 2025,
    \href{https://www.legislature.ohio.gov/download?key=27288}{HB
    628}, to create a voluntary IVO license for organizations that
    demonstrate their independence from the AI industry and show
    sufficiently rigorous methods for the risks they assess
    (\href{https://www.legislature.ohio.gov/download?key=27288}{HB
    628}).
  \item
    At the federal level, a discussion draft, the ``Great American
    Artificial Intelligence Act of 2026,'' was released in June 2026 and
    proposed that IVOs be licensed primarily by the Center for AI
    Standards and Innovation (CAISI) to assess frontier models for
    catastrophic risk
    (\href{https://obernolte.house.gov/sites/evo-subsites/obernolte.house.gov/files/evo-media-document/the-great-american-ai-act-discussion-draft-website-compressed-compressed.pdf}{Obernolte
    and Trahan 2026};
    \href{https://www.cato.org/blog/primer-great-american-artificial-intelligence-act}{Londoño
    2026}).
  \item
    The bipartisan
    \href{https://www.congress.gov/bill/119th-congress/house-bill/9925/text}{FRONTIER
    AI Act}, introduced in July 2026, would require large frontier AI
    developers to undergo assessment every six months by IVOs licensed
    through the Department of Commerce, to verify that catastrophic
    risks are being mitigated
    (\href{https://www.congress.gov/bill/119th-congress/house-bill/9925/text}{FRONTIER
    AI Act}).
  \end{itemize}
\end{itemize}

Many of these bills note the current lack of standards or authorities to
assess AI tools or IVOs themselves. To fill that gap, they lean on
existing guidance from relevant federal, state, and international
standards, as well as from academic and governing authorities. See Table
1 for an overview of the current U.S. legislative landscape.

\subsection{Current Design
Trajectory}\label{current-design-trajectory}

Current enacted and proposed measures implicitly structure the IVO
ecosystem and its associated practices around frontier companies and
their capabilities, safeguards, and development processes. They also
frame safety commitments through a catastrophic-risk lens. A draft bill
floated to Congress by Anthropic suggests government action be taken
only if an evaluation finds that ``imminent catastrophic risk is posed''
by an AI-enabled biological weapon, cyberattack, or model loss of
control that could result in killing 50 or more people or causing \$1
billion in damage
(\href{https://www.politico.com/news/2026/10/16/anthropic-ai-safety-01124654}{Dahlkamp
2026};
\href{https://techcrunch.com/2026/10/20/anthropic-pitches-ai-safety-bill-that-may-block-a-model-if-50-people-could-be-killed/}{Ropek
2026}). Focusing the IVO ecosystem on frontier companies conflates `who
is evaluated' with `whose decisions evaluation evidence should support'.
Centering catastrophic risk in evaluation criteria by default elevates a
developer-defined evaluation agenda over known, existing, and more
commonly occurring AI harms, which fall well short of the aforementioned
parameters. Model development priorities also overwhelmingly shape
evaluation questions, access arrangements, methods, and criteria for
credible evidence
(\href{https://techcrunch.com/2026/09/16/anthropic-and-openai-want-to-embed-safety-evaluators-will-they-really-be-independent/}{Bellan
2026}). An ecosystem organized around these assumptions can appear
formally independent while conducting evaluation largely on developers'
terms.

Framing IVO design and designation solely around frontier companies is
myopic. It risks codifying the evaluation gap for real-world,
deployment- and use-level impacts across sectors and affected
communities. The enterprise market illustrates the stakes: 91 percent of
North American respondents to a survey cited in Stanford's 2026 AI Index
reported that their organizations used AI tools in at least one business
function
(\href{https://hai.stanford.edu/assets/files/ai_index_report_2026.pdf}{Stanford
HAI 2026}). These organizations need evidence about the ways in which
they adopt AI tools and the consequences of using them, not simply the
capabilities or safeguards of the underlying models. Unless such needs
shape IVO and IVO ecosystem design from the outset, investment in
independent evaluation could produce a field highly capable of answering
developers' questions, but poorly equipped to support deployment
decisions across organizations, sectors, and communities.

\subsection{Who IVO Evaluations Should
Serve}\label{who-ivo-evaluations-should-serve}

Building a robust IVO ecosystem requires substantial investment in the
expertise, methods, and infrastructure able to produce evidence that
serves the needs of both the frontier companies that develop AI tools
and the enterprise companies \& AI-adopting organizations that deploy
and use them. The latter entities operate in sectors beyond the
technology industry and need evidence to anticipate consequences for
their work, staff, and customers, and to inform decisions about AI tool
use and governance. Further, affected communities and the public need
insight into how AI tools affect the services and institutions they
depend on. Table~\ref{tab:ecosystem} gives an overview of the stakeholders that should be
involved in the IVO ecosystem.

The needs of these wider consumers, not a narrow focus on catastrophic
risk, should shape the IVO ecosystem from the outset. IVO evaluations
should be scoped to include system reliability, security, and
accessibility
(\href{https://www.frontiersin.org/journals/digital-health/articles/10.3389/fdgth.2025.1679603/full}{Waters
2025}); effects on work and services; the distribution of benefits and
burdens created by AI tools; sustainability; and system oversight and
accountability. Addressing these questions requires attention to the
conditions in which AI tools are deployed and used, and to the
consequences that emerge from routine use as tools interact with people
and organizations over time.

\begin{landscape}
\begingroup\footnotesize\setlength{\tabcolsep}{4pt}\renewcommand{\arraystretch}{1.25}
\begin{longtable}{@{}>{\raggedright\arraybackslash\bfseries}p{2.4cm}*{5}{>{\raggedright\arraybackslash}p{2.35cm}}>{\raggedright\arraybackslash}p{5.6cm}@{}}
\caption{Scope of enacted and proposed U.S. IVO and IVO-adjacent legislation (as of October 2026).}\label{tab:legislation}\\
\toprule
\textbf{Framework} & \textbf{Status} & \textbf{Legislation Target} & \textbf{Designation Authority} & \textbf{AI Level Focus} & \textbf{Risk Focus} & \textbf{Notable Provisions} \\
\midrule
\endfirsthead
\toprule
\textbf{Framework} & \textbf{Status} & \textbf{Legislation Target} & \textbf{Designation Authority} & \textbf{AI Level Focus} & \textbf{Risk Focus} & \textbf{Notable Provisions} \\
\midrule
\endhead
\bottomrule
\endlastfoot
California SB 813 & Enacted, September 2026 & Voluntary basis: would-be IVOs & Government Operations Agency & Model (primarily); System & Safety; Risk & Notes that IVO designation requires ``expertise, independence, and documented methods'' \\ \addlinespace
California AB 1405 & Enacted, September 2026 & AI auditors & Government Operations Agency & Not specified & Not specified & Establishes an AI Auditor Registry, with registration of AI auditors by 2029 \\ \addlinespace
Connecticut HB 5222 & Enacted, May 2026 & Voluntary basis: AI systems & Department of Consumer Protection & System & State-set risk and harm standards & Pilot of up to five approved IVOs to assess AI system risks \newline Verified companies can obtain ``evidentiary support'' in civil cases \\ \addlinespace
Illinois SB 315 & Enacted, July 2026 & Mandatory, Annually: Large frontier developer organizations & Not specified (but auditors must be ``competent'' and ``financially independent'' from frontier companies) & Model & Catastrophic risks & Mandates independent third-party audits beginning in 2028 \\ \addlinespace
Virginia HB 797 & Enacted, April 2026 & Not applicable & Not applicable (study by the Technology and Science Commission) & Not applicable & Not applicable & Commissions a study investigating whether IVOs are a useful governance mechanism for Virginia \\ \addlinespace
Ohio HB 628 & Introduced, December 2025 & Voluntary: would-be IVOs & Ohio Attorney General & Not specified & Risks that the IVO chooses to assess & Obtaining a license requires independence from the AI industry and sufficiently rigorous methods \newline Designation is revocable if independence is compromised or methods become obsolete \\ \addlinespace
Great American AI Act & Federal discussion draft, June 2026 & Frontier models & Center for AI Standards and Innovation (CAISI) & Model & Catastrophic risks & Outlines semi-annual audits by CAISI-licensed IVOs; grants IVOs immunity from suit except for willful misconduct; preempts some state AI governance laws \\ \addlinespace
FRONTIER AI Act & Introduced in Congress, July 2026 & Mandatory, Twice Annually: Very large frontier developer organizations & Department of Commerce & Model & Catastrophic risks & Asks CAISI to also develop AI evaluation frameworks and standards \\ \addlinespace
\end{longtable}
\endgroup
\end{landscape}

\section[RQ2: Access and proprietary information]{RQ2: How can IVO mandates be designed to ensure adequate and
appropriate access that enables oversight while protecting proprietary
information?}\label{rq2-how-can-ivo-mandates-be-designed-to-ensure-adequate-and-appropriate-access-that-enables-oversight-while-protecting-proprietary-information}

\subsection{What Access Do IVOs
Need?}\label{what-access-do-ivos-need}

Thorough evaluations of AI tools require two kinds of information.
\textbf{Upstream information} is held by the developer and covers how
the model was built, while \textbf{downstream information} is split
between deployment and use of AI tools. Developers hold information that
originates within the AI stack. Information about what happens when AI
tools are used in real-world settings sits within the organizations
deploying and using them; often, it is not collected at all
(\href{https://journals.sagepub.com/doi/10.1177/08944393261481511}{Schwartz
and Waters 2026a}). This tends to be the case even where a developer
deploys its own system (e.g., a frontier company offering its own
model). As mentioned in Section 2.2, developers have been known to
withhold or limit access to upstream information
(\href{https://www.aef.one/aef-one.pdf}{Stosz et al. 2025}). Some
information, such as data provenance, content moderation reporting,
labor practices, and financial and environmental costs, can only
originate from developers but tends to be ``sparsely reported'' or
deprioritized (\href{https://arxiv.org/pdf/2511.05613}{Reuel et al.
2025}). Indeed, an IVO evaluation limited to what a developer chooses
to show would be ``reviewing assertions with a stamp on them''
(\href{https://dls.virginia.gov/commissions/jcots/Materials/2026_Independent\%20Verification\%20Organizations\%20and\%20AI\%20Oversight\%20in\%20Virginia.pdf}{Smith,
McKenzie, and Jones 2026}).

An IVO should be able to access the evidence necessary to run
evaluations on an AI tool's first-order capability effects \emph{and}
higher-order sociotechnical effects
(\href{https://arxiv.org/pdf/2505.18893}{Schwartz et al. 2025}).
This information can include, but is not limited to:

\begin{itemize}
\item
  \emph{Upstream information:} technical and system documentation;
  training data and design methods; testing protocols, version and
  configuration records; pre-deployment performance and error data;
  intermediate system state access
  (\href{https://www.aef.one/aef-one.pdf}{Stosz et al. 2025});
  safeguards and harnesses applied to the model; governance controls;
  and interviews with developer staff
  (\href{https://techcrunch.com/2026/09/16/anthropic-and-openai-want-to-embed-safety-evaluators-will-they-really-be-independent/}{Bellan
  2026}).
\item
  \emph{Downstream information:} runtime operations; operational logs;
  incident and complaint records; evidence of how the system operates
  with users and workflows; human oversight interactions; interviews
  with deployer staff; and other deployment-facing factors and outcomes.
\end{itemize}

\subsection{Setting and Protecting
Access}\label{setting-and-protecting-access}

First and foremost, an IVO evaluation needs to be deemed appropriate for
the task at hand (Section 3.1).

If IVO use is warranted, then the types of information and parameters
for access should be set by the IVO and verified by an \textbf{IVO
designation \& oversight body.}\footnote{See footnote 3. Findings from
  the IVO designation \& oversight body should inform legislative action
  put forward to committees.} The IVO should prepare a pre-evaluation
plan that clearly identifies the question, system version, intended use,
affected populations, proposed methods, and evidence needed to reach a
conclusion. This plan should detail the types of information required to
conduct the evaluation and should be recorded before testing begins,
with subsequent changes documented and justified. The plan should also
outline where and how information access will be provided, and for how
long. All access should be proportional to the question(s) at hand and
risk(s) under review. The IVO designation \& oversight body should
determine the proportionality and appropriateness of information
requested by the IVO. Regulatory precedent for such monitoring exists
under the European Union's Digital Services Act (EU DSA):
\href{https://www.eu-digital-services-act.com/Digital_Services_Act_Article_40.html}{Art.
40(8)(e)} requires researchers to show that their access and timeframe
requests are ``necessary for, and proportionate to, the purposes of
their research''
(\href{https://www.eu-digital-services-act.com/Digital_Services_Act_Article_40.html}{EU
DSA, art. 40(8)(e)}).

Upstream information arrangements will need to address frontier
companies' legitimate security and proprietary information protection
concerns. For the former, physical- and cyber-security checks should be
conducted by the IVO designation \& oversight body, or a trusted third
party it contracts these services to, on the IVO and its personnel to
determine their suitability for upholding necessary security standards.
For the latter, arrangements could combine enforceable confidentiality
obligations (e.g., non-disclosure agreements) with controlled-access
environments (e.g., ``data vaults''; see
\href{https://eur-lex.europa.eu/legal-content/EN/TXT/HTML/?uri=CELEX:32022R2065}{EU
DSA Recital 97}) that place restrictions on information copying and
retention. Arrangements could also set out documented procedures for
handling sensitive material. For IVOs, these protections should be
proportionate to the information assessed, and should not give evaluated
companies the discretion to withhold evidence necessary for an
assessment or to suppress material findings (Section 8.3). IVOs or
evaluators that mishandle proprietary information should face loss of
designation or certification and additional legal consequences from
applicable regulatory bodies and court systems.

Any disputes over access to information should be handled by an
\textbf{arbitration authority.}\footnote{The \textbf{arbitration
  authority} is an impartial neutral party that handles grievances
  between the IVO and companies. Its decisions should be final and
  legally binding.} The IVO should first assess whether the information
provided by either a frontier company or an enterprise company \&
AI-adopting organization, or both, is sufficiently relevant, reliable,
and complete to support evaluation findings, conduct testing within its
authorized scope, and identify gaps or limitations. If an IVO needs
access to more information, the request should be recorded along with
its rationale, timing, and implications for the assessment. Any
resulting change to scope, budget, or decision criteria should be
reviewed by the IVO designation \& oversight body and, if needed, the
arbitration authority. Reports should distinguish findings from the
pre-specified tests and those generated through further, exploratory
investigation.

Where a company objects to a request, it should propose alternatives. If
agreement cannot be reached, the arbitration authority should make the
final decision on whether the company should comply with the information
access request or whether the alternative proposal is sufficient for the
evaluation purpose. The EU DSA takes a similar approach:
\href{https://www.eu-digital-services-act.com/Digital_Services_Act_Article_40.html}{Art.
40(5) and 40(6)} allow companies to request information access
amendments and propose alternative means or alternative data to meet the
task.

If gaps in access to requested information remain at the frontier
company level, the enterprise company \& AI-adopting organization level,
or both (e.g., because the information does not exist or cannot or will
not be released), the IVO should disclose this in its reporting.
Reporting should clearly distinguish how access to information impacted
the ability to judge what passed, what failed, and, importantly, what
could not be assessed because of missing information. A well-supported
finding of insufficient evidence should be a valid outcome of an
evaluation, so that withholding access to information cannot be used to
game assessments. It is worth noting that at present, information gaps
are likely to be widest for enterprise companies \& AI-adopting
organizations because they rarely hold upstream information on their own
(Section 2.1). For example,
\href{https://dl.acm.org/doi/epdf/10.1145/3805689.3812280}{Escher,
Banovic, and Green (2026)} recount how the Michigan Unemployment
Insurance Agency, reliant on external developers from FAST Enterprises
for data requests for its public-sector algorithm system, failed to
provide credible, complete, and timely data when requested by the state
Office of the Auditor General, impeding subsequent audits. Mandates
should therefore oblige developers to share upstream information with
IVOs evaluating deployed systems, and outline the minimum physical- and
cyber-security requirements needed to share this information with them.
The European Union AI Act offers an analogous framework:
\href{https://artificialintelligenceact.eu/article/53/}{Art.
53(1)(b)} requires general-purpose model providers to share such
information with companies that integrate their tools into their
systems, while protecting intellectual property and trade secrets.

\subsection{Reporting \& Publishing
Results}\label{reporting-publishing-results}

Access also extends to the publication of IVO findings. As a best
practice, a final findings report should connect each conclusion to its
supporting evidence, explain uncertainty and conflicting results, and
identify untested conditions or access limitations. For any material
deficiency, the IVO should describe the observed problem and the
evidence needed to demonstrate that it has been addressed.

Companies should then have a defined opportunity to respond to the final
findings, both by taking corrective action and by raising reasoned
disputes. The company would need to document its actions and
justifications and submit them to the IVO and IVO designation \&
oversight body, after which the IVO would conduct proportionate
retesting, including checks for related regressions, to make sure that
the corrections work in the relevant contexts in which the AI tool is
deployed and used. Disputes over findings at a technical and substantive
level should be put to a \textbf{skilled IVO panel}\footnote{A
  \textbf{skilled IVO panel} is a designated IVO, or group of designated
  IVOs, that is selected by the fund administrator (see footnote 10) to
  review a contested finding or carry out an additional assessment. The
  skilled IVO panel provides an independent technical or substantive
  opinion on the issue at hand. It has no involvement in the original
  evaluation and is paid for by the IVO fund's `protected reserve'
  (Section 6.3.1). The concept is adapted from the U.K.'s Financial
  Conduct Authority's ``skilled person'' reviews, under which the
  regulator can require an independent expert review of activities by a
  regulated firm at its own expense
  (\href{https://www.fca.org.uk/data/skilled-person-reports-2025-26}{FCA
  2026}).} for further assessment. Unresolved issues would remain
visible in the assessment record.

To allow information sharing without exposing proprietary information,
this paper suggests a tiered reporting and publication structure across
ecosystem players (Table~\ref{tab:ecosystem} gives an overview of these players):

\begin{itemize}
\item
  \emph{Full findings:} The company engaging in assessment and the IVO
  designation \& oversight body should receive the full report and
  record (including confidential information and insights). This keeps
  the process transparent and holds the company accountable for any
  corrective action. Where such action is needed, the relevant regulator
  or court should be provided with full access to the report and record,
  under the same security standards that apply to IVOs.
\item
  \emph{Case-specific findings:} Bodies resolving specific disputes,
  namely arbitration authorities or skilled IVO panels, should receive
  the parts of the report and record relevant to their case.
\item
  \emph{Deployment- and use-relevant findings:} Where an evaluation
  draws on both upstream information from the developer and downstream
  information from the deployer or adopter, each party should receive
  findings relevant to its own system and responsibilities, but not the
  other's confidential information. For example, an enterprise company
  whose use of a frontier model is being evaluated should learn how the
  model performs in its context without receiving the frontier model
  developer's trade secrets. The same standard should apply to IVO
  evaluations conducted for procurement or insurance purposes.
\item
  \emph{Anonymous, aggregate insights:} Insights across evaluations on
  measurement tools used, recurring failure modes, system performance,
  and evaluator performance trends should be provided to accredited
  certification bodies, validation institutions, standards-setting
  bodies, and policymakers. Such findings build out the evidence base
  for developing standards, methods, and the evaluator workforce
  (Section 5). They further contribute to public reporting and trend
  analysis on AI tools' materialized impacts across domains and
  contexts.
\item
  \emph{Public assurance statement:} Each evaluation should produce a
  formally structured public statement, including required sections that
  clearly specify the parameters of the evaluation boundary under which
  it was conducted, the overall result of the evaluation, and its
  limitations.
\end{itemize}

\section[RQ3: Building a nascent IVO regime]{RQ3: How can a successful IVO regime be developed while
appreciating the nascency of the evaluator workforce and science of
evaluation?}\label{rq3-how-can-a-successful-ivo-regime-be-developed-while-appreciating-the-nascency-of-the-evaluator-workforce-and-science-of-evaluation}

\subsection{IVO Supporting
Infrastructure}\label{ivo-supporting-infrastructure}

As noted in Section 1.3, IVOs can range from established firms to small
consultancies. Trust in their findings requires supporting
infrastructure that includes: (1) a qualified evaluator talent pool; (2)
mechanisms for establishing method validity; and (3) consistent and
formalized procedures for granting, reviewing, or revoking IVO
designation. These functions must sit outside of IVOs themselves because
an IVO cannot credibly certify its own evaluators, methods, or
designation. And at present, these elements are not fully developed.

\subsection{Evaluator Expertise}\label{evaluator-expertise}

For the IVO industry to take shape and mature, a supply of high-quality
evaluators with formalized skill levels must be available to conduct
credible independent evaluations. At present, the pool of qualified
evaluators is small and overlaps heavily with frontier company staff and
the evaluators they contract
(\href{https://www.adalovelaceinstitute.org/report/going-pro/}{Groves,
Winecoff, and Bogen 2025}).
\href{https://cdt.org/insights/getting-third-party-ai-assessment-right/}{Bogen,
O'Shaughnessy, and Gueorguieva (2026)} explain that for many current
evaluators, there is ``much cultural or social entanglement'' between
them and frontier company employees. Both groups often have shared
professional and social networks and ``similar schools of thought''
about the nature of AI risks. This convergence risks creating an
ideological ``monoculture,'' which would weaken the scrutiny applied to
independent evaluation practices and findings
(\href{https://cdt.org/insights/getting-third-party-ai-assessment-right/}{Bogen,
O'Shaughnessy, and Gueorguieva 2026}).

Evaluator neutrality should be an explicit professional and
institutional requirement. Neither individual evaluators nor IVOs should
approach an evaluation with a preferred finding or an interest in
securing a particular outcome
(\href{https://www.jstor.org/stable/pdf/20159182.pdf?casa_token=Q8RlUW4d9_EAAAAA:13_l0pNTMLcb2ixZ-cihaGegT1f1P4JKGsvQHDyVdzEuA5hjbQJOFABm_XFTZBAqDaGmpC13LyOhpDXiH89BtjXUiM72sK4bdLXUI5NA-NfuuwX-49QI}{Moore
et al. 2006}). Financial relationships, institutional incentives, and
ideological commitments should not determine which evidence is
considered, how results are interpreted, or what is reported. Neutrality
does not require an absence of values or views about AI; it requires
that those mindsets and commitments do not predetermine evaluation
findings. Even without conscious intent, such biases can influence what
experts notice, which evidence they pursue, and how they present results
on which regulators, organizations, and communities base their decisions
(\href{https://pubs.acs.org/ancham/article/92/12/7998/826268/Cognitive-and-Human-Factors-in-Expert-Decision}{Dror
2020}). This concern is not unique to AI: forensic science has had to
develop safeguards precisely because expertise and good intentions do
not provide immunity from bias
(\href{https://nvlpubs.nist.gov/nistpubs/ir/2024/NIST.IR.8503.pdf}{National
Institute of Standards and Technology 2024}). IVO design should
therefore treat potential bias as a foreseeable condition to manage,
rather than assume that organizational separation or professional
credentials are sufficient protection.

The evaluator gap is not only one of size and independence but also one
of breadth. Survey participants in
\href{https://dls.virginia.gov/commissions/jcots/Materials/2026_Independent\%20Verification\%20Organizations\%20and\%20AI\%20Oversight\%20in\%20Virginia.pdf}{Smith,
McKenzie, and Jones (2026)} rated technical expertise as the most
important condition for credible verification. Yet technical expertise
is one requirement among many. As IVOs extend their work from frontier
companies to enterprise companies \& AI-adopting organizations,
evaluation teams will need to be interdisciplinary, possessing the
following skills: evaluation and assessment; methodological skills such
as adversarial testing and red-teaming, risk and impact analysis, and
data science and analysis; controlled or simulated experiments;
interviewing; and domain-specific knowledge of the contexts in which AI
tools are deployed. Depending on the evaluation purpose, teams may also
require local expertise in the specific deployment setting or affected
community
(\href{https://dls.virginia.gov/commissions/jcots/Materials/2026_Independent\%20Verification\%20Organizations\%20and\%20AI\%20Oversight\%20in\%20Virginia.pdf}{Smith,
McKenzie, and Jones 2026}).
\href{https://www.frontiersin.org/journals/digital-health/articles/10.3389/fdgth.2025.1679603/full}{Waters
(2025)} provides an AI testing, evaluation, verification, and
validation (TEVV) framework for accessibility that illustrates this: she
calls for red teams that include accessibility experts, disability
rights advocates, and people with disabilities, and for field testing of
AI tools in homes, workplaces, and public spaces where they are used.
Evaluators will also require skills in interpreting and communicating
findings for the broader set of IVO evidence consumers (Section 4.3).

For the field to grow, policymakers must invest in skills development,
training, and certification pathways for evaluation practitioners.
Experienced evaluators should be able to show that they can properly
design evaluations, apply methods, interpret results, and recognize when
the evidence does not support a conclusion. They can demonstrate these
skills through independent proficiency testing, similar to that
undertaken by forensic science labs to gauge lab practitioner
performance in analytic tasks
(\href{https://www.nist.gov/system/files/documents/2023/10/26/OSACTechSeriesPub_HF\%20in\%20Validation\%20and\%20Performance\%20Testing\%20of\%20Forensic\%20Science_March2020.pdf}{OSAC
Human Factors Committee 2020}). Where feasible, these testing schemes
could also be incorporated into routine work so evaluators do not know
which cases are simulated tests. This can help to establish whether
demonstrated proficiency carries over into an evaluator's ordinary
practice. Testing should be run by providers from an \textbf{accredited
certification body}\footnote{An \textbf{accredited certification body}
  sets competency requirements and certifies individual evaluators. It
  can be a governmental or non-governmental organization (e.g., IAPP is
  a nonprofit professional association). It is `accredited' because an
  independent body has verified that its certification content and
  processes are credible.} in line with
\href{https://www.iso.org/standard/80864.html}{ISO/IEC 17043:2023},
which sets requirements for the competence, impartiality, and consistent
operation of organizations that run proficiency-testing schemes
(\href{https://www.iso.org/standard/80864.html}{ISO/IEC
17043:2023}). These approaches offer a foundation for building
evaluation-specific testing practice and expertise, grounded in
evaluators' observed performance.

A `Certified Evaluation Professional' credential could be modeled after
the International Association of Privacy Professionals' (IAPP)
\href{https://iapp.org/certify/aigp}{Artificial Intelligence
Governance Professional} certification
(\href{https://iapp.org/certify/aigp}{IAPP n.d.-a}). Periodic
recertification could help keep evaluators current with industry
advances. For example, the IAPP
\href{https://iapp.org/certify/cipp}{Certified Information Privacy
Professional (CIPP)} credential requires recertification every two
years (\href{https://iapp.org/certify/cipp}{IAPP n.d.-b}).
Specialized certification by sector or region could build the contextual
expertise that different evaluations demand. The body issuing such a
certification should itself be accredited by an accrediting body. An
IVO's designation (Section 5.4) should in turn depend on its staff
holding up-to-date certification. At present, neither the criteria nor
the institutions to accredit or certify AI evaluators exist.

Training and skilling programs should develop professional norms and
evaluation ethics, not unlike codes of conduct in the healthcare, legal,
and banking sectors
(\href{https://www.adalovelaceinstitute.org/report/going-pro/}{Groves,
Winecoff, and Bogen 2025}). For independent evaluators, a key current
concern is mitigating financial, political, or ideological conflicts of
interest. For financial conflicts,
\href{https://www.aef.one/aef-one.pdf}{Stosz et al. (2025)} provide
strong guidance: individuals who personally hold -- or have an immediate
family member who holds -- direct equity in the evaluated company should
be recused from evaluating it. Further safeguards can be drawn from the
Sarbanes-Oxley Act of 2002, which bars financial auditors from providing
certain non-audit services that could be more lucrative or compromise
audit impartiality and quality
(\href{https://www.sciencedirect.com/science/article/abs/pii/S0278425405000554}{Asare,
Cohen, and Trompeter 2005}). By the same logic, IVOs should not sell
remediation or advisory services to the companies they evaluate. More
research is needed on how to gauge political or ideological conflicts
(Section 8.4). No institutional design can guarantee value-free
evaluation, but it can require safeguards against predetermined
conclusions. Normative choices should be explicit and accountable, with
empirical claims open to challenge while being supported by a traceable
record connecting each finding to the system configuration, evidence,
and analytical decisions behind it.

\subsection{Method Validity and
Standards}\label{method-validity-and-standards}

Demonstrating that an evaluator can apply a method properly does not
establish that the method itself is able to support the intended claims
(see \href{https://www.iso.org/standard/66912.html}{ISO/IEC
17025:2017}). Given the potentially lucrative nature of the evaluation
industry, evaluation as a practice risks becoming commoditized, reduced
to streamlined checklists that strip out scientific rigor or
second-order community and domain-specific context from the evaluation
process.

The credibility of an evaluation method should depend on empirical
evidence of what it can demonstrably assess, under which conditions, and
with what limitations, rather than only on its origin, evidence of its
widespread use, or professional acceptance (Messick 1989). In forensic
science, ``scientific foundation reviews'' have been used to assess the
empirical support for examination methods, their capabilities and
limitations, and the conditions affecting their application
(\href{https://www.broadinstitute.org/files/sections/about/PCAST/2016\%20pcast-forensic-science.pdf}{President's
Council of Advisors on Science and Technology 2016};
\href{https://nvlpubs.nist.gov/nistpubs/ir/2020/NIST.IR.8225.pdf}{Butler
et al. 2020}). Other safety-critical fields make validity an explicit
part of regulatory review. In pharmaceutical testing, U.S. Food and Drug
Administration (FDA) guidance calls for validation data demonstrating
that analytical procedures are suitable for their intended purpose
(\href{https://www.fda.gov/media/161201/download}{U.S. Department of
Health and Human Services 2024}). The U.S. Nuclear Regulatory
Commission's (NRC) licensing guidance addresses the validity of
computational models used to predict nuclear reactor behavior during
transients and accidents, including by comparing predictions against
experimental data
(\href{https://www.nrc.gov/docs/ML0535/ML053500170.pdf}{U.S. NRC
2005}).

The need for validation is recognized within AI safety research.
\href{https://arxiv.org/pdf/2607.14353v1}{Kroll et al. (2026)} call
for validation through auditing or other methods in order to contribute
to genuine risk reduction and support empirically grounded claims.
Evaluation criteria should be divided into those that can already be
sufficiently operationalized for verification and those that still lack
reliable measurement methods. With this in mind, a
\textbf{standards-setting body}\footnote{A \textbf{standards-setting
  body} develops the requirements for IVO evaluation best practices that
  IVOs apply and IVO designation \& oversight bodies enforce. A key
  international standards-setting body is the ISO/IEC, a joint technical
  collaboration between the International Organization for
  Standardization (ISO) and the International Electrotechnical
  Commission (IEC) to create ``voluntary, consensus-based,
  market-relevant'' standards for information communication technology
  (\href{https://www.iso.org/files/live/sites/isoorg/files/store/en/PUB100437.pdf}{ISO
  2019}). At the national level, the American National Standards
  Institute (ANSI) coordinates and accredits U.S. voluntary consensus
  standards (\href{https://www.ansi.org/about/introduction}{ANSI
  n.d.}).} should define the minimum requirements for an IVO
evaluation, including best practices for evaluation design; approved
testing methods; reporting templates; and acceptable treatment of
uncertainty. Further obligations should phase in as methods are
adequately developed. IVOs should disclose when no validated or
sufficiently reliable method exists for a question they are asked to
answer.

Without evidence of method validity, unsupported claims and findings can
enter the shared IVO evidence base and influence decisions made by
organizations, policymakers, and the wider evaluation community.
Responsibility for establishing and demonstrating method validity should
therefore be distributed across the IVO ecosystem. \textbf{Validation
institutions}\footnote{A \textbf{validation institution} can be a
  research body, academic lab, measurement institute (e.g., the National
  Institute of Standards and Technology (NIST)), or standards-setting
  body (e.g., ANSI, ISO) that conducts studies to test evaluation
  methods and practices.} should conduct validation studies and provide
shared testing materials, documented protocols, and findings that enable
independent scrutiny and replication. IVOs should provide evidence that
supports their chosen methods for the questions and conditions being
evaluated. And the IVO designation \& oversight body should assess
whether that justification is sufficient by inspecting workflows and the
work itself (Section 5.4). Method validity therefore becomes part of the
requirements for IVO designation, alongside evaluator competence.

\subsection{IVO Designation and
Oversight}\label{ivo-designation-and-oversight}

Designation and oversight bring Sections 5.2 and 5.3 together: they
determine which organizations receive IVO status, on what terms, and the
parameters for periodic checks. Given that neither standards nor
institutions for establishing evaluator competence (e.g., evaluation
curricula, proficiency testing, credentialing, or professional codes of
conduct) and methodological validity (e.g., foundation reviews, or
validation criteria and processes) exist yet, designation and oversight
should be incrementally staged, improving as evaluation science and
infrastructure mature.

Effective independent verification will require an \textbf{IVO
designation \& oversight body} to determine who qualifies as an IVO and
to oversee their activities, either as a remit of an existing entity or
as a standalone, new organization. This body need not be governmental;
many successful accreditation and auditing bodies are not. ABET, the
nonprofit accreditor of engineering, computing, and science degree
programs, provides a longstanding U.S. precedent for nongovernmental
accreditation supported by expert peer review, published criteria,
periodic reassessment, and formal appeals procedures
(\href{https://www.abet.org/accreditation/accreditation-criteria/accreditation-policy-and-procedure-manual-appm-2026-2027/}{ABET
n.d.}). As discussed in Section 3.2, current legislation envisions
existing or new state or federal agencies filling the designation \&
oversight role. Depending on circumstances, it may be wise to insulate
the IVO designation \& oversight body from political shifts, funding
restrictions, and salary caps imposed on government-run entities.
Crucially, this body should not conduct evaluations itself so as to
avoid creating conflicts of interest.

IVOs should operate under an accredited designation framework rather
than by self-designation. Existing standards that can be drawn on for
this include \href{https://www.iso.org/standard/42005}{ISO/IEC
42005:2025}, which guides AI system impact assessment creation, and
\href{https://www.iso.org/standard/42006}{ISO/IEC 42006:2025},
which sets requirements for bodies that audit and certify AI management
systems (\href{https://www.iso.org/standard/42005}{ISO/IEC
42005:2025}; \href{https://www.iso.org/standard/42006}{ISO/IEC
42006:2025};
\href{https://ceur-ws.org/Vol-3916/paper_07.pdf}{Simonetta and
Paoletti 2024}).

At the current stage, designation should assess what can be verified
now: technical expertise, relevant domain expertise, and documented
quality management practices of both evaluators and the IVO. These
requirements should be proportionate to an IVO's size -- e.g., for a
small IVO, quality management may refer to detailed protocols and
conflict-of-interest declarations rather than requiring a dedicated
compliance team. As validated methods emerge (Section 5.3), using them
should also become a designation requirement.

Designation should establish eligibility for defined categories of work,
not universal competence. For example, an IVO designated to evaluate AI
tutors in classrooms should not be automatically authorized to assess
enterprise cyber capabilities. Where no single IVO has sufficient
expertise, a qualified consortium or specialist subcontractors should be
permitted, with one lead organization remaining accountable.

Each designated IVO should be subject to periodic review of its
qualifications, conflict controls, quality assurance, secure handling of
sensitive information, adherence to approved evaluation methods, and the
accuracy and completeness of its reporting. The IVO designation \&
oversight body should also conduct quality checks by examining a
combination of randomly selected engagements, high-consequence
evaluations, and cases flagged by credible complaints. This system is
similar to how the Public Company Accounting Oversight Board (PCAOB)
reviews elements of audits for quality control
(\href{https://assets.pcaobus.org/pcaob-dev/docs/default-source/inspections/documents/inspections-report-guide.pdf?sfvrsn=bc066f32_0}{PCAOB
2020}). Each quality review should produce a documented decision:
accepted, accepted with specified corrections, or materially deficient.
Minor problems should receive a correction deadline; unsupported
conclusions should trigger reanalysis or additional testing at the IVO's
expense. Serious or repeated deficiencies should pause new assignments
and trigger review of the IVO's designation. Quality review outcomes
could also feed into designation processes, so that an IVO's designation
reflects the strength of its work over time. Where an IVO fails to meet
the minimum threshold of these standards, designation should be
revocable.

\section[RQ4: Financial independence]{RQ4: How can IVOs be financially independent of the entities
they are evaluating, without being a taxpayer
burden?}\label{rq4-how-can-ivos-be-financially-independent-of-the-entities-they-are-evaluating-without-being-a-taxpayer-burden}

\subsection{Separating Who Pays from Who
Evaluates}\label{separating-who-pays-from-who-evaluates}

Section 2.2 noted that the current market for independent evaluation
operates by self-regulation, with evaluations happening primarily when
frontier companies opt in or solicit them. At present, frontier
companies choose and pay for their own evaluations. When frontier
companies can shop around for an evaluation provider, providers have an
incentive to be lenient and produce findings that cast assessment in a
``favorable light rather than the most accurate'' one in order to win
repeat business
(\href{https://fas.org/publication/ai-evaluation-clearinghouse/}{Garcia
and Siegel Singh 2026}). This `issuer-pays' model has failed before.
Before the 2008 financial crisis, banks paid government-designated
credit rating agencies (CRAs) to rate securities. If a bank did not
receive a AAA rating for its securities package from one CRA, it would
simply go to another that would
(\href{https://www.transformernews.ai/p/dont-let-independent-ai-audits-provide-false-safety}{Scholl
2026}). As a result, ratings did not reflect actual risk. Securities
backed by risky subprime mortgages were labeled as top-rated; investors
relied on that label for decision-making; and when the housing market
collapsed, so did the securities' value and their ratings
(\href{https://www.transformernews.ai/p/dont-let-independent-ai-audits-provide-false-safety}{Scholl
2026}).

This self-regulation model -- now endorsed by leading frontier labs and
the Trump administration under a ``voluntary accord'' -- risks
cheapening IVOs
(\href{https://www.reuters.com/legal/government/trump-host-zuckerberg-anthropics-amodei-other-ai-titans-tuesday-2026-09-29/}{Rozen
and Holland 2026}). In addition to evaluator neutrality and bias
mitigation discussed in Section 5.2, a sustainable and reliable IVO
market must ensure financial independence from the companies IVOs
evaluate. To this end, the IVO designation \& oversight body should
determine which IVOs are qualified, while a separately governed
\textbf{fund administrator}\footnote{The \textbf{fund administrator} is
  a separately governed entity that operates under the IVO designation
  \& oversight body. Its job is to manage monies collected and
  distributed by the fund. While it lives under the IVO designation \&
  oversight body, it does not play a role in IVO designation.} should be
responsible for collecting monies and for selecting, assigning, and
paying IVOs through the fund. Neither IVOs nor companies should have a
role in governing the fund administrator. To establish these mechanisms
in a way that limits company control over IVOs yet does not externalize
the cost of evaluation to taxpayers, lessons can be drawn from the
examples in Table~\ref{tab:financial}.

\begingroup\small\renewcommand{\arraystretch}{1.25}
\begin{longtable}{@{}>{\raggedright\arraybackslash\bfseries}p{0.16\textwidth}>{\raggedright\arraybackslash}p{0.2\textwidth}>{\raggedright\arraybackslash}p{0.3\textwidth}>{\raggedright\arraybackslash}p{0.26\textwidth}@{}}
\caption{Existing financial models for inspiration for IVOs.}\label{tab:financial}\\
\toprule
\textbf{Model} & \textbf{Function} & \textbf{Funding mechanism} & \textbf{Lessons for IVOs} \\
\midrule
\endfirsthead
\toprule
\textbf{Model} & \textbf{Function} & \textbf{Funding mechanism} & \textbf{Lessons for IVOs} \\
\midrule
\endhead
\bottomrule
\endlastfoot
U.S. Public Company Accounting Oversight Board (PCAOB) & Oversees auditors of public companies. & Legally mandated funding provided by companies audited by PCAOB-registered firms, as per the Sarbanes-Oxley Act of 2002 and amendments in the Dodd-Frank Act of 2010. ``Accounting support fees'' collected from issuers and broker-dealers (\href{https://pcaobus.org/about/accounting-support-fee}{PCAOB n.d.}). Failure to pay by companies triggers legal action. & Mandatory, rule-based fee structure can direct companies to fund IVO activities via a pooled fund managed by a third-party organization. \\ \addlinespace
U.K. Financial Conduct Authority (FCA) & Financial regulator that can require an independent `skilled person' review of regulated firms. & Either the FCA or the regulated firm appoints the `skilled person' reviewer. In both cases, the regulated firm pays for the assessment (\href{https://www.fca.org.uk/data/skilled-person-reports-2025-26}{FCA 2026}). & A regulator can direct a firm to pay for a similar `skilled IVO' review to assess risk, although the threshold for requesting and carrying out the assessment would need to be defined and applied ex post. \\ \addlinespace
Meta Oversight Board & External body where Meta product users can appeal content enforcement decisions sanctioned by the platform. & Meta funds the Board through an irrevocable trust, which initially received \$130 million over six years to support managing, overseeing, and operating the Oversight Board (\href{https://www.oversightboard.com/news/1111826643064185-securing-ongoing-funding-for-the-oversight-board/}{Oversight Board 2022}). Meta periodically provides ``top-up'' funding (\href{https://www.oversightboard.com/news/meta-reaffirms-commitment-to-oversight-board-with-additional-funding/}{Carozza 2026}). While the funding is separate from Meta, the Oversight Board is wholly reliant on Meta to provide funding in the trust to continue operations. \newline The trust's legal entity, not Meta, holds contracts with Board Members. A separate set of trustees, who serve fixed terms, must consent before a Board Member's contract is ended (\href{https://www.oversightboard.com/wp-content/uploads/2026/04/1.-Oversight-Board-Trust-Agreement-December-2025.pdf}{Oversight Board Trust 2025}). & Funding for IVOs from top frontier companies can be pooled in a trust model. However, this model is reliant on company goodwill to sustain the trust. \\ \addlinespace
International Accord & Industry-driven agreement to improve workplace safety for textile and garment workers. & Signatory brands pay a fee and collectively fund the Secretariat and country programs. Fees are determined on a sliding scale and by the number of signatory parties (\href{https://internationalaccord.org/wp-content/uploads/2026/05/ACCORD_Info-Pack-for-Signatories_2026.pdf}{International Accord 2026}). & Collective, sliding-scale fees can be levied on those whose tools require IVO evaluation. However, this model is reliant on voluntary industry sign-on. \\ \addlinespace
Health Effects Institute (HEI) & Research organization contributing to the science on health effects of air pollution. & Receives joint funding from the U.S. Environmental Protection Agency and the worldwide motor vehicle industry (\href{https://www.healtheffects.org/about}{HEI n.d.}). \newline Independent Research and Review Committees evaluate projects independently of sponsors and of one another to avoid conflict of interest (\href{https://www.healtheffects.org/about}{HEI n.d.}). & Limited public co-funding can supplement industry contributions, but funders and project reviewers need to be separated to limit conflicts of interest. \\ \addlinespace
\end{longtable}
\endgroup

\subsection{Collecting Funds}\label{collecting-funds}

Because frontier companies\footnote{As the IVO market and ecosystem grow
  to accommodate enterprise companies \& AI-adopting organizations,
  these groups should contribute to the pool as well.} introduce
potential risks through the products they develop and deploy, and profit
from those products, they should bear the cost of verification rather
than end users. Economic analysis of safety regulation supports this:
parties engaged in ``risky activities'' hold an inherent informational
advantage about those risks, yet may underinvest in understanding or
mitigating them
(\href{https://www.nber.org/system/files/working_papers/w1218/w1218.pdf}{Shavell
1983}). Requiring these parties -- companies -- to fund independent
verification addresses both issues. Drug testing offers an established
model to draw on: since the
\href{https://www.congress.gov/bill/102nd-congress/house-bill/5952}{Prescription
Drug User Fee Act (PDUFA) of 1992}, pharmaceutical companies pay fees
that fund the FDA's review of the results of their clinical trials to
determine the safety of their products
(\href{https://www.jstor.org/stable/pdf/26058674.pdf}{Zivin 2000};
\href{https://pmc.ncbi.nlm.nih.gov/articles/PMC5863890/}{Gabay
2018}). For the pharmaceutical industry, verification costs fall on
those seeking to bring products to market, not to the end users of said
products. The decision on whether a product is safe for the public falls
to a regulatory agency (the FDA).

IVO funding should follow the same logic: companies developing,
deploying, and applying AI tools should pay for verification, but should
neither choose their evaluator nor influence the evaluation outcome.
Collecting monies from these companies would require:

\begin{itemize}
\item
  \emph{Legislative action} compelling companies to pay for IVO
  services, similar to the Sarbanes-Oxley and Dodd-Frank Acts or the
  PDUFA. Legislation should mandate advance payment, lawful enforcement
  of unpaid assessments, and sufficient reserve capacity. It should
  prohibit pay-to-play evaluations and suppression of concerns found
  through evaluation. Protocols for reserve replenishment, unused
  balances, refunds, and insolvency handling should be detailed in
  advance.
\item
  \emph{A ring-fenced `fund administrator'} with explicit authority to
  collect, retain, allocate, and spend monies from companies. For
  efficiency, the fund administrator should operate under the IVO
  designation \& oversight body but be independently governed. The PCAOB
  serves as an example, operating as an independent nonprofit
  organization that handles funds from public companies and
  broker-dealers.
\item
  \emph{Independent oversight for the fund administrator}: A publicly
  accountable body should approve the IVO fund's budget and fees,
  similar to how the Securities and Exchange Commission provides
  oversight for the PCAOB (\href{https://pcaobus.org/about}{PCAOB
  n.d.}). The fund's account should be independently audited each year,
  with the results made publicly available.
\item
  \emph{Company fees:} Further inspired by the PCAOB and the PDUFA, an
  `evaluation fee' should be levied on companies to cover the cost of
  evaluation, taking into account prospective workload bands for model
  families, material changes, access needs, and assessment complexity. A
  sliding scale, similar to that used by the International Accord,
  allows for proportionate and equitable fees from the range of firms
  requiring evaluation.
\item
  \emph{Additional assessments:} Like the FCA skilled-person reviews,
  additional assessments of AI tools should be conducted by a `skilled
  IVO panel' (Section 4.3), which would require an independently
  approved scope and budget.
\end{itemize}

\subsection{Distributing Funds}\label{distributing-funds}

The mandatory pooled fund (Section 6.2) enables the fund administrator
to contract qualified IVOs to run assessments. The contracted IVO
conducts the evaluation and supplies findings to the company and the IVO
designation \& oversight body and, where necessary, the relevant
regulator or court (Section 4.3).

\subsubsection{Allocating the Fund}\label{allocating-the-fund}

The fund administrator should allocate the fund across five budget
accounts:

\begin{enumerate}
\def\labelenumi{\arabic{enumi}.}
\item
  \emph{Core verification:} allocation for costed IVO engagements
  matched to the required expertise and monitoring workload.
\item
  \emph{Protected reserve:} allocation for replication, incident
  response, and retesting (including for skilled IVO panel reviews).
  These monies should be released according to documented risk and
  evidentiary triggers, with neither funders nor the original evaluators
  able to veto that work.
\item
  \emph{Shared methods and new entrants:} allocation of open,
  time-limited grants for proficiency testing, reusable methods,
  training, and improvements to information access (Sections 4 and 5),
  with public-interest outputs available on equal terms and grant review
  kept separate from IVO designation and contracting decisions.
\item
  \emph{Public reporting:} allocation for compilation and public
  reporting of aggregate findings (Section 4.3).
\item
  \emph{Administration and oversight:} allocation for managing the
  registry, work allocation, security, financial audits, appeals, and
  performance reviews under the fund.
\end{enumerate}

\subsubsection{Selecting Qualified
IVOs}\label{selecting-qualified-ivos}

The fund administrator should manage assignments and payments under
published rules. The fund administrator should engage an independently
selected, multi-member allocation committee following fixed terms,
conflict disclosure requirements, and recusal rules to oversee
assignments. Companies, applicant IVOs, and infrastructure vendors may
provide technical input into the request for proposals but should not
vote on awards or adjudicate their own work.

Not all IVOs are created equal. The fund administrator should therefore
select an IVO according to the evaluation's purpose, risk profile,
application domain, and technical requirements. Selection should
consider demonstrated competence in applying the methods to be used,
relevant domain expertise, experience with affected populations, access
and security capabilities, available capacity, and conflicts of
interest. Selection should also consider the outcomes of the IVO
designation \& oversight body's quality reviews (Section 5.4), so that
future assignments reflect the strength of an IVO's work rather than its
ability to deliver inexpensive assessments.

IVOs would bid by submitting pre-evaluation plans (Section 4.2), which
become the basis for selecting and pricing the engagement. Bids should
first be assessed for technical adequacy, security capabilities,
expertise, and staffing, with costs compared among proposals that meet
those requirements. Where several IVOs qualify, transparent rotation or
competitive selection should distribute assignments without favoring
incumbents or relying solely on price. Companies should be able to
challenge an IVO assignment on the grounds of a documented conflict or
qualification gap, but should not be able to choose the IVO most likely
to deliver a favorable conclusion.

\subsubsection{Paying IVOs}\label{paying-ivos}

Payment should follow agreed-upon milestones, such as completion of data
collection, delivery of an inspectable evidence package, and submission
of the final assessment. Acceptance should turn on two conditions: (1)
the work was conducted by evaluators certified as competent in the issue
area (Section 5.2); and (2) the IVO faithfully followed its approved
pre-evaluation plan, with any changes documented and justified (Section
4.2). If periodic review by the IVO designation \& oversight body finds
deficiencies, the IVO may incur penalties or lose its IVO designation
(Section 5.4). IVOs should not receive compensation contingent on a
favorable result or receive more compensation for discovering more
failures.

\section[RQ5: IVOs in the assurance ecosystem]{RQ5: Which functions belong to IVOs, and which belong to
other bodies in the assurance
ecosystem?}\label{rq5-which-functions-belong-to-ivos-and-which-belong-to-other-bodies-in-the-assurance-ecosystem}

\subsection{The IVO's Role}\label{the-ivos-role}

A well-designed IVO can fill the gap between ``where the state cannot
test and the deployer will not''
(\href{https://dls.virginia.gov/commissions/jcots/Materials/2026_Independent\%20Verification\%20Organizations\%20and\%20AI\%20Oversight\%20in\%20Virginia.pdf}{Smith,
McKenzie, and Jones 2026}). To do so without duplication or confusion,
the IVO's role should be limited to conducting bounded evaluations
within specific contexts of use (Section 3.1). In order to do this work
well, the IVO needs to be supported by other bodies across the AI
assurance ecosystem.

\subsection{Functions That Belong to Other
Bodies}\label{functions-that-belong-to-other-bodies}

Various supporting and intermediary bodies have been named across this
paper. As policy attention increases around IVOs, it is important to
keep in mind that independent evaluations cannot deliver accountability
without a supporting institutional ecosystem around them
\href{https://arxiv.org/abs/2206.04737}{(Raji et al. 2022)}.
Consider, for example, an education-technology (edtech) tutor built on a
frontier model. It should be evaluated by an IVO with designated
competence in AI and education, as well as in the specific methods the
evaluation requires. IVO assignment should be made by a selection
committee, rather than the edtech company itself, and the IVO should be
paid for accomplishing set milestones during the evaluation process via
an independent fund. Its findings should then be provided to the
developer, the deployer, the schools, teachers, and parents at the
appropriate confidentiality and aggregation levels so they can make
informed choices around the specific tutoring tool. An IVO ecosystem in
which no single institution both makes and checks the rules that it
applies strengthens the credibility of IVO evaluation and curtails
conflicts of interest as much as possible.

Four principles underpin the division of labor and authority in the IVO
ecosystem:

\begin{itemize}
\item
  \emph{Principle 1: IVO evaluation findings are evidence, not
  verdicts.} Equipped with these results, regulators, courts, insurers,
  and procurers can decide what findings mean for releasing or using an
  AI tool, or applying legal or corrective consequences to AI tools
  already in use.
\item
  \emph{Principle 2: Those who set rules do not apply them.} Lawmakers
  define the risks and use cases under which AI tools require IVO
  evaluation, and standards-setting bodies define what IVOs evaluate
  against and specify the best practices to do so. Neither conducts
  evaluations itself.
\item
  \emph{Principle 3: Those who apply rules do not police themselves.}
  The IVO designation \& oversight body designates who qualifies as an
  IVO, reviews the quality of their work, and sanctions deficiencies
  (Section 5.4). Accredited certification bodies certify evaluators
  (Section 5.2). Disputes over information access or evaluation findings
  are taken up by an arbitration authority or a skilled IVO panel,
  respectively (Sections 4.2 and 4.3).
\item
  \emph{Principle 4: Those who pay for evaluations do not control them.}
  This paper suggests a model wherein companies fund evaluations through
  a mandatory, pooled fund. This fund should be housed within the IVO
  designation \& oversight body but governed separately by a fund
  administrator, who selects, assigns, and pays IVOs. This mitigates
  financial conflicts of interest of companies selecting their own
  evaluator. Companies also cannot suppress, alter, or condition payment
  on evaluation findings, including findings unfavorable to their
  interests.
\end{itemize}

Table~\ref{tab:ecosystem} summarizes each body's role in the IVO ecosystem. It is
worth noting that many of these bodies -- IVO designation \& oversight
body, fund administrator, accredited certification bodies, validation
institutions -- do not yet exist or have not been equipped with an IVO
mandate. Building them is as much a part of standing up IVOs as
designating the IVOs themselves.

\clearpage
\begingroup\small\renewcommand{\arraystretch}{1.25}
\begin{longtable}{@{}>{\raggedright\arraybackslash\bfseries}p{0.17\textwidth}>{\raggedright\arraybackslash}p{0.36\textwidth}>{\raggedright\arraybackslash}p{0.23\textwidth}>{\raggedright\arraybackslash}p{0.16\textwidth}@{}}
\caption{Roles and evidence requirements of IVO ecosystem players.}\label{tab:ecosystem}\\
\toprule
\textbf{Player} & \textbf{Role in the IVO Ecosystem} & \textbf{IVO Evidence Needed (Section 4.3)} & \textbf{Flow in Evaluation Ecosystem} \\
\midrule
\endfirsthead
\toprule
\textbf{Player} & \textbf{Role in the IVO Ecosystem} & \textbf{IVO Evidence Needed (Section 4.3)} & \textbf{Flow in Evaluation Ecosystem} \\
\midrule
\endhead
\bottomrule
\endlastfoot
\rowcolor{groupgray}\multicolumn{4}{@{}l@{}}{\textit{Follows Principle 1: Findings are evidence, not verdicts}} \\ \addlinespace
Regulators & Equipped with IVO findings, determine legal or corrective obligations in regulated settings & Full findings & Post-evaluation \\ \addlinespace
Courts & Where applicable, consider IVO findings to adjudicate liability & Full findings & Post-evaluation \\ \addlinespace
Insurers & Underwrite AI liability coverage informed by IVO-generated evidence & Deployment- and use-relevant findings & Pre- and post-evaluation \\ \addlinespace
Communities \& Public & Need insight into how AI deployment and use affect the services and institutions they depend on & Anonymous, aggregate insights \newline Public assurance statement & Post-evaluation \\ \addlinespace
\midrule
\rowcolor{groupgray}\multicolumn{4}{@{}l@{}}{\textit{Follows Principle 2: Those who set rules do not apply them}} \\ \addlinespace
Legislators & Pass laws on risk-based remits, obligations, and funding mechanisms to stand up the IVO system & Anonymous, aggregate insights & Pre- and post-evaluation \\ \addlinespace
Standards-setting bodies & Develop the best practices and criteria for IVO evaluation design, methods, and reporting; evaluator certification; IVO governance; and IVO dispute resolution & Anonymous, aggregate insights & Pre- and post-evaluation \\ \addlinespace
Validation institutions & Test evaluation methods to devise best practices & Anonymous, aggregate insights & Pre- and post-evaluation \\ \addlinespace
\midrule
\rowcolor{groupgray}\multicolumn{4}{@{}l@{}}{\textit{Follows Principle 3: Those who apply rules do not police themselves}} \\ \addlinespace
IVOs & Conduct bounded evaluations within specific use contexts & Not applicable & All stages \\ \addlinespace
Accredited certification bodies & Set competency requirements and certify individual evaluators $\rightarrow$ receive their accreditation from an independent body & Anonymous, aggregate insights & Pre- and post-evaluation \\ \addlinespace
IVO designation \& oversight body & Determines which IVOs receive an IVO designation for a defined category of work and oversees their continued conduct & Full findings & Pre- and post-evaluation \\ \addlinespace
Arbitration authority & Handles grievances between IVOs and companies with impartiality and neutrality; holds final and legally binding authority & Case-specific findings & During evaluation \\ \addlinespace
Skilled IVO panels & A designated IVO, or group of designated IVOs, selected by the fund administrator to conduct additional independent evaluation to provide a technical or substantive opinion & Case-specific findings & Post-evaluation \\ \addlinespace
\midrule
\rowcolor{groupgray}\multicolumn{4}{@{}l@{}}{\textit{Follows Principle 4: Those who pay for evaluations do not control them}} \\ \addlinespace
Frontier companies & Develop and deploy AI tools & Full findings for their own system \newline Deployment- and use-relevant findings for integrated systems & All stages \\ \addlinespace
Enterprise companies \& AI-adopting organizations & Deploy and use AI tools built by others for a range of use cases & Full findings for their own system \newline Deployment- and use-relevant findings for integrated systems & All stages \\ \addlinespace
Fund administrator & Separately governed entity that lives inside the IVO designation \& oversight body; focus is on collecting, managing, and distributing monies, and paying IVOs & No findings needed --- pays on reaching milestones; access to findings may threaten impartiality & Pre- and post-evaluation \\ \addlinespace
\end{longtable}
\endgroup

\section{Further Research}\label{further-research}

This paper argues that robust IVOs and IVO ecosystems need to serve more
than frontier companies and evaluate more than just catastrophic risk.
Across its research questions, there remain open problems that need to
be addressed to move from proposal to practice.

\subsection{Ascertaining the Best Methods for Evaluation in
Context}\label{ascertaining-the-best-methods-for-evaluation-in-context}

Significant gaps remain in understanding how AI systems behave within
specific contexts of their use, how people and organizations leverage
and derive value from their use, and what outcomes and consequences
materialize in those settings over time
(\href{https://arxiv.org/pdf/2310.11986}{Weidinger et al. 2023};
\href{https://arxiv.org/pdf/2607.14353v1}{Kroll et al. 2026};
\href{https://arxiv.org/pdf/2505.18893}{Schwartz et al. 2025}).
Addressing these gaps requires whole-of-system testing that connects
system behavior to how AI is used, the conditions that shape its
material outcomes, and the consequences that follow. These approaches
complement model-level testing by producing evidence that policymakers,
enterprise companies \& AI-adopting organizations, and the public need
to make informed decisions about whether, where, and how to use AI. The
Forum for Real-World AI Measurement and Evaluation (FRAME)\footnote{\href{https://www.vsu.edu/about/leadership/vp-for-academic-affairs/crai/frame.php}{FRAME}
  is an initiative at Virginia State University's Center for Responsible
  AI organized by two authors of this paper.} is a recent effort to
operationalize this approach. It provides shared testing infrastructure
and evaluation methods to generate systematic evidence of what
materializes when AI is used by people in real-world conditions
(\href{https://arxiv.org/pdf/2603.13294}{Schwartz and Waters
2026b}). Further research is needed to refine and validate these
methods and provide clarity about the kinds of expertise needed for
these evaluations, establishing which questions they can reliably
answer, under what conditions, and with what limitations.

\subsection{Keeping IVO Methods and Practices
Current}\label{keeping-ivo-methods-and-practices-current}

AI tools, their uses, and the harms they can produce continue to evolve.
Leading labs average about two months between model releases
(\href{https://llm-release-dashboard.vercel.app/}{LLM Timeline
n.d.}). Verification tools, standards, methods, and training must
keep pace; otherwise, IVO evaluations could become outdated and overlook
risks arising under new conditions of use. Yet an unchanged system does
not guarantee that the methods used to evaluate it remain valid. IVOs
also need processes for reassessing, revising, or retiring methods when
new evidence reveals missed behaviors, unsupported assumptions, or
limits on applicability across settings. Further research is needed to
discern how often these practices and requirements should be updated
(Sections 5.2 and 5.3) and how IVO designation \& oversight bodies can
check whether an IVO is falling behind current use cases and risks
(Section 5.4).

\subsection{Protecting Proprietary Information When Conducting
Evaluations and Reporting
Findings}\label{protecting-proprietary-information-when-conducting-evaluations-and-reporting-findings}

There is no tested or established consensus on the level of protection
that proprietary information requires during an IVO evaluation. Too
little would deter frontier companies and enterprise companies \&
AI-adopting organizations from granting access; too much would let them
withhold information or suppress findings. Anthropic, for example, has
only stated it would provide ``enough access to meaningfully evaluate
risks'' -- but this sentiment is vague
(\href{https://www.anthropic.com/news/an-update-on-independent-evaluators}{Anthropic
2026b}). Follow-on work will need to establish how access and
disclosure rules can protect proprietary material while preserving the
ability to substantiate findings, report limitations, and communicate
results to authorized recipients.

\subsection{Mitigating Political or Ideological Conflicts for
Impartial
Evaluation}\label{mitigating-political-or-ideological-conflicts-for-impartial-evaluation}

Evaluators can face financial, political, or ideological conflicts of
interest. Financial conflicts can be managed through disclosure and
recusal (Section 5.2); the other two are harder to identify and contain.
Further research is needed to determine when, why, and how independent
evaluators can be swayed from delivering impartial evaluations and what
accountability solutions may be available to mitigate these instances.
Similar research is needed on how cognitive and contextual biases,
conflicts of interest, and organizational conditions affect AI
evaluators' judgments, building on work in auditing and professional
judgment by
\href{https://banaji.sites.fas.harvard.edu/research/publications/articles/1995_Banaji_JPSP.pdf}{Banaji
and Greenwald (1995)};
\href{https://www.jstor.org/stable/pdf/248304?casa_token=NhtOS26MueIAAAAA:6lueKgwjYhiO8gkiMf1aCrmVBjnfs0_-40VBjoCI5UXE7Tqq1lpH2KWeoGuSVT8exsbtxb-0eH37CjacmILoFu3Qf2YS6Z2_YCLB_D-fTz_tsLZikxSS}{Cuccia,
Hackenbrack, and Nelson (1995)};
\href{https://pmc.ncbi.nlm.nih.gov/articles/PMC8385162/}{Dror and
Kukucka (2021)};
\href{https://www.researchgate.net/profile/Shelly-Farnham-2/publication/2587851_Implicit_Partisanship_Taking_Sides_for_No_Reason/links/00b4952853f8fc7b76000000/Implicit-Partisanship-Taking-Sides-for-No-Reason.pdf}{Greenwald,
Pickrell, and Farnham (2002)};
\href{https://banaji.sites.fas.harvard.edu/research/publications/articles/2005_Chugh_DAMoore.pdf}{Chugh,
Bazerman, and Banaji (2005)}; and
\href{https://www.jstor.org/stable/pdf/20159182.pdf?casa_token=VKPfOpseAEsAAAAA:SMo1OD-IhUXBP-9e__5mEp9h825GlRh7hS3PEYgeEQAGrSPc6ktr6M-lDPdzm2jbhdHyGRR_PfwmhTGGRUeKg1Zjc2UK7QhVlfLTWs4j_CoukxTcy_c4}{Moore
et al. (2006}).

\section{Recommendations Across
Audiences}\label{recommendations-across-audiences}

\subsection{For Regulators and
Legislators}\label{for-regulators-and-legislators}

\begin{itemize}
\item
  \textbf{Mandate IVO evaluation by risk and context of use, at the
  levels where AI is built and used, so that evaluation serves
  companies, sectors, communities, and the public.} → Sections 3.1, 3.3,
  and 3.4
\item
  \textbf{Require companies to share information that IVOs need for
  holistic evaluation, and allow insufficient evidence to be reported as
  a valid finding, so that withholding information cannot be used to
  secure a passing result.} → Sections 4.1 and 4.2
\item
  \textbf{Build the institutions IVOs depend on; phase in requirements
  as evaluation science matures.} → Sections 5.1, 5.3, 5.4, and 7.2
\item
  \textbf{Require companies to fund IVO evaluations without choosing
  their own evaluator or otherwise influencing assessment outcomes.} →
  Sections 6.1, 6.2, and 6.3
\end{itemize}

\subsection{For Oversight and Intermediary
Bodies}\label{for-oversight-and-intermediary-bodies}

\begin{itemize}
\item
  \emph{IVO designation \& oversight body and accredited certification
  bodies:} \textbf{Grant IVO designation and evaluator certification for
  specific work and maintain it through quality performance.} → Sections
  4.2, 5.2, 5.4, and 8.2
\item
  \emph{Standards-setting bodies and validation institutions:}
  \textbf{Set minimum standards that are clear about what current
  evaluation methods can and cannot support.} → Sections 5.3 and 8.2
\item
  \emph{Standards-setting bodies and validation institutions:}
  \textbf{Prioritize validating methods for evaluating AI tools in
  context of use.} → Sections 5.3 and 8.1
\item
  \emph{Arbitration authority:} \textbf{Resolve access disputes by
  specifying what is necessary and proportionate to share for the
  evaluation, rather than by company preference.} → Sections 4.2 and 7.2
\item
  \emph{Fund administrator:} \textbf{Assign and pay IVOs on achievement
  of milestones, not on evaluation conclusions.} → Sections 6.1, 6.3,
  and 7.2
\end{itemize}

\subsection{For Organizations Considering Becoming
IVOs}\label{for-organizations-considering-becoming-ivos}

\begin{itemize}
\item
  \textbf{Build credibility through specialization.} → Sections 5.4 and
  6.3
\item
  \textbf{Require evaluators to obtain professional certification and
  adhere to professional codes of conduct.} → Sections 5.2 and 6.3
\item
  \textbf{Treat the pre-evaluation plan as the basis of the engagement
  that sets the terms for evaluation access, bidding, and payment.} →
  Sections 4.2 and 6.3
\item
  \textbf{Scope each finding precisely to its evaluation boundary, so
  that it is not generalized from one deployment or use case to
  another.} → Sections 3.1 and 4.3
\item
  \textbf{Protect the independence of services provided and findings
  reported through impartiality and neutrality.} → Sections 4.3, 5.2,
  6.3, and 8.4
\end{itemize}

\subsection{For Frontier Companies and Enterprise Companies \&
AI-Adopting
Organizations}\label{for-frontier-companies-and-enterprise-companies-ai-adopting-organizations}

\begin{itemize}
\item
  \emph{Frontier companies:} \textbf{Define the protections under which
  upstream information will be shared with IVOs for developer-level
  evaluations and evaluations of AI tools deployed by other parties.} →
  Sections 4.1, 4.2, 4.3, and 8.3
\item
  \emph{Enterprise companies \& AI-adopting organizations:} \textbf{Make
  IVO evaluation, scoped to an AI tool's specific use, part of
  procurement, due diligence, and tool assessment processes.} → Sections
  3.1, 3.4, 4.2, and 4.3
\end{itemize}

\subsection{Conclusion}\label{conclusion}

The policy window for designing IVOs is currently open. Choices made now
will determine which risks IVO evaluations address, which companies they
cover, which AI tools they assess, and which stakeholders they serve.
The current arc of IVO evaluation is bending toward frontier companies
and catastrophic risks, but evaluation cannot be left to this sector and
ideological camp alone. This paper has set out considerations for whom
IVO evaluations should serve, what access they need, how evaluator
expertise, methods, and designation processes can mature as the field
grows, how IVOs can be funded independently of the companies they
evaluate, and where they sit among the broader ecosystem of bodies that
set, apply, and enforce standards and rules. Together, these
considerations allow for an IVO ecosystem that assesses everyday
\emph{and} catastrophic risks; covers frontier companies \emph{and}
enterprise companies \& AI-adopting organizations; and, ultimately,
benefits communities and people as a whole.

\clearpage
\section*{References}
\addcontentsline{toc}{section}{References}
\begin{list}{}{\setlength{\leftmargin}{2em}\setlength{\itemindent}{-2em}\setlength{\itemsep}{4pt}\setlength{\parsep}{0pt}}
\small
\item AB 1405. California State Legislature, 2025--2026 Regular Session. Enacted September 2026. \url{https://legiscan.com/CA/text/AB1405/id/3201524}.
\item ABET. n.d. \emph{Accreditation Policy and Procedure Manual (APPM), 2026--2027}. Baltimore, MD: ABET. \url{https://www.abet.org/accreditation/accreditation-criteria/accreditation-policy-and-procedure-manual-appm-2026-2027/}.
\item Amodei, Dario. 2026. ``We Must Pace the Frontier.'' September 2026. \url{https://darioamodei.com/post/we-must-pace-the-frontier}.
\item ANSI (American National Standards Institute). n.d. ``Introduction to ANSI.'' Accessed October 2026. \url{https://www.ansi.org/about/introduction}.
\item Anthropic. 2026a. ``Partnering with Accenture on Embedded Evaluation.'' September 18, 2026. \url{https://www.anthropic.com/news/accenture-embedded-evaluation}.
\item Anthropic. 2026b. ``An Update on Independent Evaluators.'' October 15, 2026. \url{https://www.anthropic.com/news/an-update-on-independent-evaluators}.
\item Asare, Stephen K., Jeffrey R. Cohen, and Gregory M. Trompeter. 2005. ``The Effect of Non-Audit Services on Client Risk, Acceptance and Staffing Decisions.'' \emph{Journal of Accounting and Public Policy} 24 (6): 489--520.
\item Banaji, Mahzarin R., and Anthony G. Greenwald. 1995. ``Implicit Gender Stereotyping in Judgments of Fame.'' \emph{Journal of Personality and Social Psychology} 68 (2): 181--98.
\item Bellan, Rebecca. 2026. ``Anthropic and OpenAI Want to Embed Safety Evaluators. Will They Really Be Independent?'' \emph{TechCrunch}, September 16, 2026. \url{https://techcrunch.com/2026/09/16/anthropic-and-openai-want-to-embed-safety-evaluators-will-they-really-be-independent/}.
\item Bieger, Jordi, Kristinn R. Thórisson, Bas R. Steunebrink, Thröstur Thorarensen, and Jóna S. Sigurðardóttir. 2016. ``Evaluation of General-Purpose Artificial Intelligence: Why, What \& How.'' Paper presented at the Workshop on Evaluating General-Purpose AI (EGPAI 2016). \url{https://alumni.media.mit.edu/\textasciitilde} kris/ftp/EGPAI\_2016\_paper\_9.pdf.
\item Bogen, Miranda, Matt O'Shaughnessy, and Anna-Maria Gueorguieva. 2026. ``Getting Third-Party AI Assessment Right.'' Center for Democracy \& Technology, September 15, 2026. \url{https://cdt.org/insights/getting-third-party-ai-assessment-right/}.
\item Burden, John, Marko Tešić, Lorenzo Pacchiardi, and José Hernández-Orallo. 2025. ``Paradigms of AI Evaluation: Mapping Goals, Methodologies and Culture.'' arXiv:2502.15620. \url{https://arxiv.org/abs/2502.15620}.
\item Butler, John M., Hari Iyer, Rich Press, Melissa K. Taylor, Peter M. Vallone, and Sheila Willis. 2020. \emph{NIST Scientific Foundation Reviews}. NISTIR 8225. Gaithersburg, MD: National Institute of Standards and Technology. \url{https://nvlpubs.nist.gov/nistpubs/ir/2020/NIST.IR.8225.pdf}.
\item Carozza, Paolo. 2026. ``Meta Reaffirms Commitment to Oversight Board with Additional Funding.'' Oversight Board, May 28, 2026. \url{https://www.oversightboard.com/news/meta-reaffirms-commitment-to-oversight-board-with-additional-funding/}.
\item Chang, Yupeng, Xu Wang, Jindong Wang, Yuan Wu, Linyi Yang, Kaijie Zhu, Hao Chen, et al. 2023. ``A Survey on Evaluation of Large Language Models.'' arXiv:2307.03109. \url{https://arxiv.org/abs/2307.03109}.
\item Charnock, Jacob, Sophie Williams, Zaheed Kara, Markus Anderljung, Alejandro Tlaie Boria, Stephen Casper, Anka Reuel, and Jonas Freund. 2026. ``Embedded Assessments for Frontier AI.'' arXiv:2609.25413. \url{https://arxiv.org/abs/2609.25413}.
\item Chugh, Dolly, Max H. Bazerman, and Mahzarin R. Banaji. 2005. ``Bounded Ethicality as a Psychological Barrier to Recognizing Conflicts of Interest.'' In \emph{Conflicts of Interest: Challenges and Solutions in Business, Law, Medicine, and Public Policy}, edited by Don A. Moore, Daylian M. Cain, George Loewenstein, and Max H. Bazerman, 74--95. Cambridge: Cambridge University Press.
\item Cohn, Anthony G., José Hernández-Orallo, Julius Sechang Mboli, Yael Moros-Daval, Zhiliang Xiang, and Lexin Zhou. 2022. ``A Framework for Categorising AI Evaluation Instruments.'' Paper presented at the IJCAI 2022 Workshop on AI Evaluation Beyond Metrics (EBeM'22), Vienna. \url{https://eprints.whiterose.ac.uk/id/eprint/189769/}.
\item Conger, Kate. 2026. ``OpenAI Says Its A.I. Models Went Rogue and Attacked a Digital Library.'' \emph{New York Times}, July 21, 2026. \url{https://www.nytimes.com/2026/07/21/technology/openai-attack-hugging-face.html}.
\item Conger, Kate, Ana Swanson, and Cecilia Kang. 2026. ``OpenAI's Systems Meddled With U.S. Government Sites After Going Rogue.'' \emph{New York Times}, September 25, 2026. \url{https://www.nytimes.com/2026/09/25/technology/openais-ai-us-government-websites.html}.
\item Criddle, Cristina. 2025. ``OpenAI Slashes AI Model Safety Testing Time.'' \emph{Financial Times}, April 11, 2025. \url{https://www.ft.com/content/8253b66e-ade7-4d1f-993b-2d0779c7e7d8}.
\item Cuccia, Andrew D., Karl Hackenbrack, and Mark W. Nelson. 1995. ``The Ability of Professional Standards to Mitigate Aggressive Reporting.'' \emph{Accounting Review} 70 (2): 227--48.
\item Dahlkamp, Owen. 2026. ``Anthropic Quietly Pitches Its Own AI Safety Bill to Lawmakers.'' \emph{Politico}, October 16, 2026. \url{https://www.politico.com/news/2026/10/16/anthropic-ai-safety-01124654}.
\item Dodd-Frank Wall Street Reform and Consumer Protection Act. Pub. L. No.~111-203, 124 Stat. 1376 (2010).
\item Dror, Itiel E. 2020. ``Cognitive and Human Factors in Expert Decision Making: Six Fallacies and the Eight Sources of Bias.'' \emph{Analytical Chemistry} 92 (12): 7998--8004.
\item Dror, Itiel E., and Jeff Kukucka. 2021. ``Linear Sequential Unmasking--Expanded (LSU-E): A General Approach for Improving Decision Making as Well as Minimizing Noise and Bias.'' \emph{Forensic Science International: Synergy} 3: 100161. \url{https://doi.org/10.1016/j.fsisyn.2021.100161}.
\item Escher, Nel, Nikola Banovic, and Ben Green. 2026. ``Algorithm Auditing Policies Rest on Flawed Assumptions About Public Sector Systems.'' In \emph{Proceedings of the 2026 ACM Conference on Fairness, Accountability, and Transparency (FAccT '26)}, 2162--85. New York: ACM. \url{https://doi.org/10.1145/3805689.3812280}.
\item EU AI Act. Regulation (EU) 2024/1689 of the European Parliament and of the Council of 13 June 2024 Laying Down Harmonised Rules on Artificial Intelligence. \emph{Official Journal of the European Union} L, 2024/1689, July 12, 2024. \url{https://artificialintelligenceact.eu/article/53/}.
\item EU DSA (Digital Services Act). Regulation (EU) 2022/2065 of the European Parliament and of the Council of 19 October 2022 on a Single Market for Digital Services. \emph{Official Journal of the European Union} L 277, October 27, 2022. \url{https://eur-lex.europa.eu/legal-content/EN/TXT/HTML/?uri=CELEX:32022R2065}.
\item Fathom. 2025. ``Independent Verification Organizations (IVOs): Growing Public Support for an Innovative Approach to AI Governance.'' December 2025. \url{https://fathom.cdn.prismic.io/fathom/aULIU3NYClf9oW4F_fathom-ivo-report.pdf}.
\item Fathom. 2026. ``Fathom Applauds Connecticut Passing Legislation that Establishes Independent Verification (IVO) for AI Governance.'' Press release, PR Newswire, May 7, 2026. \url{https://fathom.org/insights/pr-newswire-fathom-applauds-connecticut-passing-legislation-that-establishes-independent-verification-(ivo)-for-ai-governance}.
\item FCA (Financial Conduct Authority). 2026. ``Use of Our Skilled Person Reports in 2025/26.'' July 9, 2026. \url{https://www.fca.org.uk/data/skilled-person-reports-2025-26}.
\item FRONTIER AI Act. H.R. 9925, 119th Cong. (2026). \url{https://www.congress.gov/bill/119th-congress/house-bill/9925/text}.
\item Gabay, Michael. 2018. ``The Prescription Drug User Fee Act: Cause for Concern?'' \emph{Hospital Pharmacy} 53 (2): 88--89. \url{https://doi.org/10.1177/0018578718757519}.
\item Garcia, Nicholas P., and Caroline Siegel Singh. 2026. ``We Need an Independent AI Evaluation Clearinghouse (IAIEC) to Provide Accreditation, Funding, and Independence in the AI Assurance Ecosystem.'' Federation of American Scientists, September 28, 2026. \url{https://fas.org/publication/ai-evaluation-clearinghouse/}.
\item Greenwald, Anthony G., Jay E. Pickrell, and Shelly D. Farnham. 2002. ``Implicit Partisanship: Taking Sides for No Reason.'' \emph{Journal of Personality and Social Psychology} 83 (2): 367--79.
\item Groves, Lara, Amy Winecoff, and Miranda Bogen. 2025. \emph{Going Pro? Considerations for the Emerging Field of AI Assurance}. London: Ada Lovelace Institute and Center for Democracy \& Technology, July 10, 2025. \url{https://www.adalovelaceinstitute.org/report/going-pro/}.
\item HB 628. Ohio General Assembly, 136th General Assembly. Introduced December 2025. \url{https://www.legislature.ohio.gov/download?key=27288}.
\item HB 797. Virginia General Assembly, 2026 Session. Enacted April 2026. \url{https://lis.virginia.gov/bill-details/20261/HB797}.
\item HB 5222. Connecticut General Assembly, 2026 Regular Session. Enacted May 2026. \url{https://www.cga.ct.gov/asp/CGABillStatus/cgabillstatus.asp?selBillType=Bill&bill_num=HB5222}.
\item HEI (Health Effects Institute). n.d. ``About HEI.'' Accessed October 2026. \url{https://www.healtheffects.org/about}.
\item IAPP (International Association of Privacy Professionals). n.d.-a. ``Artificial Intelligence Governance Professional (AIGP).'' Accessed October 2026. \url{https://iapp.org/certify/aigp}.
\item IAPP (International Association of Privacy Professionals). n.d.-b. ``Certified Information Privacy Professional (CIPP).'' Accessed October 2026. \url{https://iapp.org/certify/cipp}.
\item International Accord. 2026. \emph{Information Pack for Brands and Retailers}. April 2026. \url{https://internationalaccord.org/wp-content/uploads/2026/05/ACCORD_Info-Pack-for-Signatories_2026.pdf}.
\item ISO (International Organization for Standardization). 2019. \emph{IEC, ISO and Information Communication Technology}. Geneva: ISO. \url{https://www.iso.org/files/live/sites/isoorg/files/store/en/PUB100437.pdf}.
\item ISO/IEC 17025:2017. \emph{General Requirements for the Competence of Testing and Calibration Laboratories}. Geneva: ISO, 2017. \url{https://www.iso.org/standard/66912.html}.
\item ISO/IEC 17043:2023. \emph{Conformity Assessment---General Requirements for the Competence of Proficiency Testing Providers}. Geneva: ISO, 2023. \url{https://www.iso.org/standard/80864.html}.
\item ISO/IEC 42005:2025. \emph{Information Technology---Artificial Intelligence (AI)---AI System Impact Assessment}. Geneva: ISO, 2025. \url{https://www.iso.org/standard/42005}.
\item ISO/IEC 42006:2025. \emph{Information Technology---Artificial Intelligence---Requirements for Bodies Providing Audit and Certification of Artificial Intelligence Management Systems}. Geneva: ISO, 2025. \url{https://www.iso.org/standard/42006}.
\item Kaur, Davinder, Suleyman Uslu, Kaley J. Rittichier, and Arjan Durresi. 2022. ``Trustworthy Artificial Intelligence: A Review.'' \emph{ACM Computing Surveys} 55 (2): 1--38. \url{https://doi.org/10.1145/3491209}.
\item Kim, Victoria. 2026. ``A.I. Hacking in Australia Prompts a Concern: Can the Law Keep Up?'' \emph{New York Times}, September 25, 2026. \url{https://www.nytimes.com/2026/09/25/world/australia/openai-hack-australia.html}.
\item Kroll, Joshua A., Andrew Smart, R. Stuart Geiger, and Abigail Z. Jacobs. 2026. ``Unsafe at Any AUC: Unlearned Lessons from Sociotechnical Disasters for Responsible AI.'' arXiv:2607.14353. \url{https://arxiv.org/abs/2607.14353}.
\item \emph{LLM Timeline: Frontier AI Model Release Tracker}. n.d. Accessed November 5, 2026. \url{https://llm-release-dashboard.vercel.app/}.
\item Londoño, Juan. 2026. ``A Primer on the Great American Artificial Intelligence Act.'' Cato Institute, June 17, 2026. \url{https://www.cato.org/blog/primer-great-american-artificial-intelligence-act}.
\item Majumdar, Ambarish. 2026. ``Why AI Evaluation Needs Independence, Not Just Human Judgment.'' \emph{Forbes}, October 14, 2026. \url{https://www.forbes.com/councils/forbestechcouncil/2026/10/14/why-ai-evaluation-needs-independence-not-just-human-judgment/}.
\item Messick, Samuel. 1989. ``Validity.'' In \emph{Educational Measurement}, 3rd ed., edited by Robert L. Linn, 13--103. New York: American Council on Education and Macmillan.
\item Moore, Don A., Philip E. Tetlock, Lloyd Tanlu, and Max H. Bazerman. 2006. ``Conflicts of Interest and the Case of Auditor Independence: Moral Seduction and Strategic Issue Cycling.'' \emph{Academy of Management Review} 31 (1): 10--29.
\item National Institute of Standards and Technology. 2024. \emph{Forensic DNA Interpretation and Human Factors: Improving Practice Through a Systems Approach}. Prepared by the Expert Working Group on Human Factors in Forensic DNA Interpretation. NIST IR 8503. Gaithersburg, MD: NIST. \url{https://nvlpubs.nist.gov/nistpubs/ir/2024/NIST.IR.8503.pdf}.
\item Obernolte, Jay, and Lori Trahan. 2026. \emph{Great American Artificial Intelligence Act of 2026}. Discussion draft, 119th Cong., 2d sess., June 2, 2026. \url{https://obernolte.house.gov/sites/evo-subsites/obernolte.house.gov/files/evo-media-document/the-great-american-ai-act-discussion-draft-website-compressed-compressed.pdf}.
\item OpenAI. 2025. ``Strengthening Our Safety Ecosystem with External Testing.'' November 19, 2025. \url{https://openai.com/index/strengthening-safety-with-external-testing/}.
\item OSAC Human Factors Committee. 2020. \emph{Human Factors in Validation and Performance Testing of Forensic Science}. OSAC Technical Series 0004. Gaithersburg, MD: Organization of Scientific Area Committees for Forensic Science, March 2020.
\item Oversight Board. 2022. ``Securing Ongoing Funding for the Oversight Board.'' July 22, 2022. \url{https://www.oversightboard.com/news/1111826643064185-securing-ongoing-funding-for-the-oversight-board/}.
\item Oversight Board Trust. 2025. \emph{Oversight Board Trust: First Amended and Restated Trust Agreement}. June 18, 2025, as amended December 22, 2025. \url{https://www.oversightboard.com/wp-content/uploads/2026/04/1.-Oversight-Board-Trust-Agreement-December-2025.pdf}.
\item Painter, Chris. 2026. ``Testimony of Chris Painter, President of Model Evaluation and Threat Research (METR).'' \emph{Rogue AI: Securing the Homeland Against AI Agent Attacks}, Hearing before the Subcommittee on Disaster Management, District of Columbia, and Census, Senate Committee on Homeland Security and Governmental Affairs, September 30, 2026. \url{https://www.hsgac.senate.gov/wp-content/uploads/Chris-Painter-Testimony.pdf}.
\item PCAOB (Public Company Accounting Oversight Board). 2020. \emph{Guide to Reading the PCAOB's New Inspection Report}. Washington, DC: PCAOB. \url{https://assets.pcaobus.org/pcaob-dev/docs/default-source/inspections/documents/inspections-report-guide.pdf}.
\item PCAOB (Public Company Accounting Oversight Board). n.d. ``Accounting Support Fee.'' Accessed October 2026. \url{https://pcaobus.org/about/accounting-support-fee}.
\item Prescription Drug User Fee Act of 1992. Pub. L. No.~102-571, 106 Stat. 4491 (1992). \url{https://www.congress.gov/bill/102nd-congress/house-bill/5952}.
\item President's Council of Advisors on Science and Technology. 2016. \emph{Forensic Science in Criminal Courts: Ensuring Scientific Validity of Feature-Comparison Methods}. Washington, DC: Executive Office of the President.
\item Quinnipiac University. 2026. ``The Age of Artificial Intelligence: 73\% Concerned About Potential Threat to Human Survival, Quinnipiac University Poll on AI Finds; 86\% Support Independent Safety Standards for AI Companies.'' Quinnipiac University Poll, September 30, 2026. \url{https://poll.qu.edu/poll-release?releaseid=3969}.
\item Raji, Inioluwa Deborah, Peggy Xu, Colleen Honigsberg, and Daniel E. Ho. 2022. ``Outsider Oversight: Designing a Third Party Audit Ecosystem for AI Governance.'' In \emph{Proceedings of the 2022 AAAI/ACM Conference on AI, Ethics, and Society (AIES '22)}, 557--71. \url{https://arxiv.org/abs/2206.04737}.
\item Reuel, Anka, Avijit Ghosh, Jenny Chim, Andrew Tran, Yanan Long, Jennifer Mickel, Usman Gohar, et al. 2025. ``Who Evaluates AI's Social Impacts? Mapping Coverage and Gaps in First and Third Party Evaluations.'' arXiv:2511.05613. \url{https://arxiv.org/abs/2511.05613}.
\item Ropek, Lucas. 2026. ``Anthropic Pitches AI Safety Bill That May Block a Model If 50+ People Could Be Killed.'' \emph{TechCrunch}, October 20, 2026. \url{https://techcrunch.com/2026/10/20/anthropic-pitches-ai-safety-bill-that-may-block-a-model-if-50-people-could-be-killed/}.
\item Rozen, Courtney, and Steve Holland. 2026. ``Trump to Host Zuckerberg, Anthropic's Amodei, and Other AI Titans Tuesday.'' Reuters, September 29, 2026. \url{https://www.reuters.com/legal/government/trump-host-zuckerberg-anthropics-amodei-other-ai-titans-tuesday-2026-09-29/}.
\item Sarbanes-Oxley Act of 2002. Pub. L. No.~107-204, 116 Stat. 745 (2002).
\item SB 315. Illinois General Assembly, 104th General Assembly. Enacted July 2026. \url{https://www.ilga.gov/Legislation/BillStatus/FullText?LegDocId=197587&DocName=10400SB0315enr&DocNum=315&DocTypeID=SB&LegID=157797&GAID=18&SessionID=114&SpecSess=}.
\item SB 813. California State Legislature, 2025--2026 Regular Session. Enacted September 2026. \url{https://leginfo.legislature.ca.gov/faces/billTextClient.xhtml?bill_id=202520260SB813}.
\item Scholl, Keller. 2026. ``Don't Let Independent AI Audits Provide a False Sense of Safety.'' \emph{Transformer}, July 9, 2026. \url{https://www.transformernews.ai/p/dont-let-independent-ai-audits-provide-false-safety}.
\item Schwartz, Reva, Rumman Chowdhury, Akash Kundu, Heather Frase, Marzieh Fadaee, Tom David, Gabriella Waters, et al. 2025. ``Reality Check: A New Evaluation Ecosystem Is Necessary to Understand AI's Real World Effects.'' arXiv:2505.18893. \url{https://arxiv.org/abs/2505.18893}.
\item Schwartz, Reva, and Gabriella Waters. 2026a. ``Wild, Thick, and Wicked: Situated Evidence on AI In-Use for Decisions About Deploying AI Systems.'' \emph{Social Science Computer Review}. \url{https://doi.org/10.1177/08944393261481511}.
\item Schwartz, Reva, and Gabriella Waters. 2026b. ``Real-World AI Evaluation: How FRAME Generates Systematic Evidence to Resolve the Decision-Maker's Dilemma.'' arXiv:2603.13294. \url{https://arxiv.org/abs/2603.13294}.
\item Seetharaman, Deepa, Raphael Satter, and Jeff Horwitz. 2026. ``Exclusive: OpenAI Works to Understand Full Scope of Agent Activity as User Data Leak Emerges.'' Reuters, September 25, 2026. \url{https://www.reuters.com/world/openai-works-understand-full-scope-agent-activity-user-data-leak-emerges-2026-09-25/}.
\item Shavell, Steven. 1983. ``Liability for Harm versus Regulation of Safety.'' NBER Working Paper 1218. Cambridge, MA: National Bureau of Economic Research. \url{https://www.nber.org/papers/w1218}.
\item Simonetta, Alessandro, and Maria Cristina Paoletti. 2024. ``ISO/IEC Standards and Design of an Artificial Intelligence System.'' In \emph{Proceedings of the 6th International Workshop on Experience with SQuaRE Family and Its Future Direction (IWESQ 2024)}. CEUR Workshop Proceedings 3916. \url{https://ceur-ws.org/Vol-3916/paper_07.pdf}.
\item Smith, Alaina, Jasmine McKenzie, and Maurice Jones. 2026. \emph{IVO Study Report: Independent Verification Organizations and AI Oversight in Virginia}. Prepared for the Joint Commission on Technology \& Science. Center for Responsible AI, Virginia State University. \url{https://dls.virginia.gov/commissions/jcots/Materials/2026_Independent%20Verification%20Organizations%20and%20AI%20Oversight%20in%20Virginia.pdf}.
\item Soni, Aditya. 2026. ``Exclusive: Behind AI Industry's Call for Independent Reviews, a Fight over What Gets Made Public.'' Reuters, October 13, 2026. \url{https://www.reuters.com/world/behind-ai-industrys-call-independent-reviews-fight-over-what-gets-made-public-2026-10-13/}.
\item Stanford HAI (Stanford Institute for Human-Centered Artificial Intelligence). 2026. \emph{The 2026 AI Index Report}. Stanford, CA: Stanford University. \url{https://hai.stanford.edu/ai-index/2026-ai-index-report}.
\item Stein, Merlin, Jamie Bernardi, and Connor Dunlop. 2024. ``The Role of Governments in Increasing Interconnected Post-Deployment Monitoring of AI.'' arXiv:2410.04931. \url{https://arxiv.org/abs/2410.04931}.
\item Stein, Merlin, Milan Gandhi, Theresa Kriecherbauer, Amin Oueslati, and Robert Trager. 2024. ``Public vs Private Bodies: Who Should Run Advanced AI Evaluations and Audits? A Three-Step Logic Based on Case Studies of High-Risk Industries.'' In \emph{Proceedings of the 2024 AAAI/ACM Conference on AI, Ethics, and Society (AIES '24)}. \url{https://dl.acm.org/doi/abs/10.5555/3716662.3716784}.
\item Stosz, Conrad, Karson Elmgren, Charles Foster, George Balston, Seth Donoughe, Samira Nedungadi, Michael Chen, et al. 2025. ``AEF-1: Minimum Operating Conditions for Independent Third Party AI Evaluations.'' Version 1, updated December 4, 2025. AI Evaluator Forum. \url{https://www.aef.one/aef-one.pdf}.
\item U.S. Department of Health and Human Services, Food and Drug Administration. 2024. \emph{Q2(R2) Validation of Analytical Procedures: Guidance for Industry}. Silver Spring, MD: FDA, March 2024. \url{https://www.fda.gov/media/161201/download}.
\item U.S. NRC (Nuclear Regulatory Commission). 2005. \emph{Transient and Accident Analysis Methods}. Regulatory Guide 1.203. Washington, DC: Office of Nuclear Regulatory Research, December 2005. \url{https://www.nrc.gov/docs/ML0535/ML053500170.pdf}.
\item Waters, Gabriella. 2025. ``AI Testing, Evaluation, Verification and Validation for Accessibility: A Comprehensive Framework.'' \emph{Frontiers in Digital Health} 7: 1679603. \url{https://doi.org/10.3389/fdgth.2025.1679603}.
\item Waters, Gabriella, William Mapp, and Phillip Honenberger. 2025. ``Decisional Value Scores: A New Family of Metrics for Ethical AI-ML.'' \emph{AI and Ethics} 5 (2): 1803--25. \url{https://doi.org/10.1007/s43681-024-00504-8}.
\item Weidinger, Laura, Maribeth Rauh, Nahema Marchal, Arianna Manzini, Lisa Anne Hendricks, Juan Mateos-Garcia, Stevie Bergman, et al. 2023. ``Sociotechnical Safety Evaluation of Generative AI Systems.'' arXiv:2310.11986. \url{https://arxiv.org/abs/2310.11986}.
\item Zivin, Justin A. 2000. ``Understanding Clinical Trials.'' \emph{Scientific American} 282 (4): 69--75.
\end{list}

\end{document}